\documentclass{article}
\pdfoutput=1
\usepackage{arxiv}

\usepackage{url}
\usepackage[english]{babel}
\usepackage{numprint}
\usepackage{rotating}
\usepackage{natbib}
\usepackage{enumerate}

\usepackage{tikz}
\usetikzlibrary{trees,positioning,shapes,shadows,arrows}
\tikzset{
  basic/.style  = {draw, text width=3.2cm, font=\sffamily, rectangle},
  root/.style   = {basic, rounded corners=2pt, thin, drop shadow, align=center, fill=white, font=\large},
  level-2/.style = {basic, rounded corners=4pt, thin, drop shadow, align=center, fill=white, font=\large, text width=1.2cm},
  level-3/.style = {basic,  rounded corners=6pt,  thin, align=center, fill=white, text width=1.5cm, font=\small},
  level-4/.style = {basic, thin, align=center, fill=white, text width=1.5cm, font=\footnotesize}
}

\newcommand{\htt}{\textsf{HTT}}
\newcommand{\ctt}{\textsf{CTT}}
\newcommand{\ett}{\textsf{ETT}}

\newcommand{\uett}{\textsf{UETT}}
\newcommand{\cbctt}{\textsf{CB-CTT}}
\newcommand{\pectt}{\textsf{PE-CTT}}
\newcommand{\itcii}{\textsf{ITC-2007-ETT}}
\newcommand{\xhstt}{\textsf{XHSTT}}
\newcommand{\itciv}{\textsf{ITC-2019}}

\newcommand{\opthub}{\textsc{OptHub}}

\begin{document}

\title{Educational Timetabling: Problems, Benchmarks, and State-of-the-Art Results}

\author{Sara Ceschia, Luca Di Gaspero, Andrea Schaerf\thanks{Corresponding author}\\
DPIA, University of Udine, Via delle Scienze 206, 33100 Udine, Italy\\
email: \url{{sara.ceschia,luca.digaspero,andrea.schaerf}@uniud.it}
}

\maketitle
\begin{abstract}
  We propose a survey of the research contributions on the field of
  Educational Timetabling with a specific focus on ``standard''
  formulations and the corresponding benchmark instances. We identify
  six of such formulations and we discuss their features, pointing out
  their relevance and usability.  Other available formulations and
  datasets are also reviewed and briefly discussed. Subsequently, we report the
  main state-of-the-art results on the selected benchmarks, in terms of
  solution quality (upper and lower bounds), search techniques,
  running times, statistical distributions, and other side settings.
\end{abstract}

\keywords{Timetabling \and Validation \and Benchmarks\and Reproducibility}

\section{Introduction}
\label{sec:intro}

Educational Timetabling, in essence, consists in assigning
teacher/student meetings to days, timeslots, and classrooms. Despite this
apparent simplicity, experience teaches us that every single institution
has its own rules, conventions, and fixations, thus making each
specific problem almost unique.  As a consequence, uncountably many
different problem formulations have been proposed in the literature on
Educational Timetabling, depending on the type of institution
(high-school, university, or other), the type of meetings (lectures,
exams, \dots), and the different settings, constraints, and
objectives.

Many papers in the literature tackle a specific problem using a
selected search method.  The authors normally claim the success of the
application, though rarely dispelling the doubt over the
readers that the method used was more the authors' ``favorite''
rather than the most suitable for the problem under consideration.
A few previous surveys have tried to put in order this situation by creating a
taxonomy of both problem formulations and corresponding search
methods used for their solution, in order to draw some conclusions
about what works best in each specific case (see Section~\ref{sec:timetabling}). 

In this survey, we want to take a somewhat different point of view.
Specifically, we focus on the review of the problem formulations and
their publicly available datasets, critically discussing their
practical relevance and usability. To this aim, we highlight which
datasets have been considered most frequently in the literature, so
that they have risen to the status of \emph{benchmarks}, and the
corresponding formulation to the status of a \emph{de facto} standard.  

We identified six standard formulations, which are
presented in chronological order in
Sections~\ref{sec:uett}---\ref{sec:itciv}. Incidentally, the
chronological order corresponds also to the order of increasing
complexity and adherence to the real-world situation. Indeed, we can
see that the research has moved continuously from very simplified
problems toward full-fledged ones. Nonetheless, in our opinion the
early simplified formulations are still interesting testbeds for new
search methods, and they have not yet finished to serve their purpose. On
the contrary, the accumulated bulk of results and techniques make them
even more interesting and challenging.

For these formulations and benchmarks, we discuss state-of-the-art
results, in terms of solution quality, search techniques, running
times, and other side settings. We will also discuss the availability
of upper and lower bounds, in order to identify which are the most
challenging instances for future comparisons.

We also review and discuss other formulations that have not attracted
general interest so far, but still provide real-world publicly
available datasets and could be potentially interesting for the
community.

Finally, we consider the issue of reliability of the results claimed
in the literature, stressing the importance of the presence of
instance and solution checkers, so as to provide against possible
errors and misunderstandings. To this aim, we developed a web
application, named \opthub{} (\url{https://opthub.uniud.it}), that allows 
users to check and upload both new instances and solutions. All data, properly
validated and timestamped, is available for download and inspection,
along with scoreboards and statistics.  The system is meant to provide
a unified and up-to-date site for current contributions, so as to
facilitate and encourage further research and future comparisons.
\opthub{}, whose development is still ongoing, currently hosts four of
the formulations discussed in this survey.  The formulations hosted
are the early ones that do not have a dedicated and updated online
repository on their own.

In a way, this survey is meant for researchers interested in writing
what \citet{John02} called a \emph{horse race paper}, in which the
authors assess the quality of their methods by the comparison to previous
research on the designated benchmarks. We aim to help such perspective
researchers to be rigorous, fair, and comprehensive as much as
possible in such a complex task of comparing with the whole literature.

However, our hope is that this effort could be useful also
for the authors of an \emph{application paper} (still following
\citeauthor{John02}'s terminology), that aims at solving one specific
original problem. Indeed, those authors could evaluate the quality of
their search method by identifying an underlying standard problem that
could be a simplified version of their own specific one, adapt their
search method to solve it, and report the corresponding results.
Naturally, it is not expected that a solver for a complex,
full-fledged problem could outperform specialized ones for the
benchmarks, but this would give a reasonable measure of the quality of
the proposed approach.

This survey is organized as follows. In Section~\ref{sec:timetabling}, we
list the various problems within the scope of the Educational Timetabling
area. In Section~\ref{sec:formulations-benchmarks}, we introduce and
discuss the available formulations and datasets for these problems. In
Section~\ref{sec:results}, we illustrate and comment the
state-of-the-art results for the benchmarks.  Finally, conclusions and
future directions are discussed in Section~\ref{sec:conclusions}.

\section{Educational Timetabling}
\label{sec:timetabling}

In this section, we introduce the educational timetabling problems and
discuss various general issues of the research area.

\subsection{Educational timetabling problems}
\label{sec:problems}

According to the literature on timetabling \citep[see,
e.g.,][]{Scha99,King13}, there are three main problems in the
educational timetabling area:
\begin{description}
\item[High-School Timetabling (\htt)] The weekly scheduling for all the
  classes of a high-school, avoiding teachers meeting two classes at the same
  time, and vice versa.
\item[University Course Timetabling (\ctt)] The weekly scheduling for
  all the lectures of a set of university courses, minimizing the
  overlaps of lectures of courses having common students.
\item[University Examination Timetabling (\ett)] The scheduling for
  the exams of a set of university courses, avoiding overlap of exams
  of courses having common students, and spreading the exams for the
  students as much as possible.
\end{description}

Even though a clear cut between \htt, \ctt, and \ett{} is not
possible (e.g., some high-schools are organized in a university
fashion), they normally differ from each other significantly, and most
of the papers in the literature can be classified within one of
these three problems.

\subsection{Previous surveys}
Many surveys on educational timetabling have recently appeared 
in the literature. However, due to
the vastness of the research area, all of them focus an a subset of
the problems introduced in the previous section, in order to reduce
their scope.  For example, \citet{BuPe02} and \citet{MiHa13} focus on university
timetabling (\ctt{} and \ett), likewise \citet{Lewi08} who further limits
his study to metaheuristic techniques.  Similarly, the survey by
\citet{QBMML09} is dedicated to \ett, whereas the one by
\citet{Pill14} is only on \htt, and the recent ones by
\citet{CSGSK21} and \citet{TGKS21} are on \ctt{} and \htt,
respectively. The survey by \citet{BCRT15} reviews only one specific
formulation of \ctt, namely the curriculum-based course timetabling
(\cbctt), that will be introduced and discussed in
Section~\ref{sec:cb-ctt}.

\subsection{Other timetabling problems}
There are also other problems within the Educational Timetabling field
that have been addressed in the literature, although they are less
popular than the previous three. Among these ``minor'' problems we can
include \emph{Student Sectioning} \citep{MuMu10}, \emph{Thesis Defense
  Timetabling} \citep{BCDDS19}, \emph{Trainee/Intern/Resident
  Assignment} (for medical and military schools) \citep{AkMa21}, and
\emph{Conference Scheduling} \citep{StPV18}.  We do not discuss the
above problems in details, as there are no available datasets that
have reached the status of benchmarks.  An exception is Student
Sectioning that is included together with \ctt{} in the \itciv{}
formulation, that will be discussed in Section~\ref{sec:itciv}.

Other timetabling problems, which fall outside the scope of
Educational Timetabling, such as Employee Timetabling \citep{MeSc03},
Transportation (trains and airplanes) Timetabling \citep{CaTo12}, and
Sport Timetabling \citep{VGSG20} are not discussed here.

\subsection{Timetabling initiatives}
The timetabling community is quite active. There are a biannual
conference series (\url{http://patatconference.org}) and a EURO
Working Group (\url{https://www.euro-online.org/web/ewg/14/}), both
called PATAT (Practice and Theory of Automated Timetabling) and
dedicated to the whole area of Timetabling problems. One of their
activities has been the organization of five International Timetabling
Competitions: ITC-2002, ITC-2007 \citep{MSPM10}, ITC-2011
\citep{PDKMS2013}, ITC-2019 \citep{MuRM18}, and ITC-2021
\citep{VGBD21}. These competitions have brought forth most of the
standard formulations and benchmarks discussed in
Section~\ref{sec:formulations-benchmarks}. Incidentally, the most
recent one, ITC-2021, did not focus on educational timetabling like
the previous ones but on sports timetabling.

\subsection{Multiobjective formulations}
For all the standard formulations that we will introduce in
Section~\ref{sec:formulations-benchmarks}, there is a single objective
function, defined as a weighted sum of the various penalty terms to be
minimized. Therefore, we do not include in this survey the issues
related to multiobjective optimization \citep{SiBP04}, although the
multiobjective perspective would be surely useful in this context, as
objectives in timetabling could be rather intangible and thus not
always commensurable. Indeed, many objectives are related to the
comfort of the participants (students or teachers), so that it is
difficult to assign to them a specific numeric weight.  Furthermore,
besides the classical objectives measuring the general comfort, some
authors include also other notions, which are even more difficult to
be put in the same scale of the other objectives. These include the
\emph{fairness} \citep{MuWa16}, that takes care for the balanced
distribution of the discomfort among the participants (teachers and
students) and the \emph{robustness} \citep{AkGu18}, that measures the
possibility to do not deteriorate the quality in presence of
unforeseen disruptions.

\subsection{Terminology and taxonomy}

We define here some common terms in the timetabling vocabulary that
will be used throughout this survey. Concepts that are specific of one
formulation are introduced in the dedicated section.

\begin{description}
\item[Times:] The time \emph{horizon} is divided into \emph{days} and
  each day is split into \emph{timeslots} (in general, the same number
  of timeslots is given in each day). A \emph{period} is a pair
  $\langle \textrm{day}, \textrm{timeslot} \rangle$.
\item[Events:] An \emph{event} is a meeting between students and one
  or more teachers.  Events can be of different types: \emph{lectures}
  or \emph{exams} of a \emph{course}, \emph{laboratories}, or
  \emph{seminars}. 
\item[Resources:] We consider three main kinds of resources: students,
  teachers, and rooms. Events have to be scheduled taking into account
  resource restrictions, such as \emph{students}' enrollments,
  \emph{teachers}' requests and \emph{rooms}' availabilities.
\item[Constraints:] As customary, constraints are split into
  \emph{hard} and \emph{soft} ones (soft constraints are also called
  objectives).  The hard constraints must be always satisfied, whereas
  the soft ones contribute to the objective function, which is a
  weighted sum of all soft constraint penalties.
\end{description}

\begin{figure}
\begin{small}
    \centering
\begin{tikzpicture}[
  level 1/.style={sibling distance=15em, level distance=5em},
  level 2/.style={sibling distance=6em, level distance=5em},
  edge from parent/.style={->,solid,black,thick,sloped,draw}, 
  edge from parent path={(\tikzparentnode.south) -- (\tikzchildnode.north)},
  >=latex, node distance=1.2cm, edge from parent fork down]

\node[root] {\textbf{Educational Timetabling}}
  child {node[level-2] (c3) {\textbf{\htt}}
      child {node[level-3] (c31) {\xhstt}}
  }
  child {node[level-2] (c1) {\textbf{\ctt}}
      child {node[level-3] (c11) {\cbctt}}
      child {node[level-3] (c12) {\pectt}}
      child {node[level-3] (c13) {\itciv}}
  }
  child {node[level-2] (c2) {\textbf{\ett}}
      child {node[level-3] (c21) {\uett}}
      child {node[level-3] (c22) {\itcii}}
  };

 \begin{scope}[every node/.style={level-4}]
 \node [below of = c11, xshift=10pt] (c111) {\textsf{comp}};
 \node [below of = c12, xshift=10pt] (c121) {ITC-2002};
 \node [below of = c121] (c122) {ITC-2007};
 \node [below of = c13, xshift=10pt] (c131) {ITC-2019};
 \node [below of = c21, xshift=10pt] (c211) {Toronto};
 \node [below of = c22, xshift=10pt] (c221) {ITC-2007};
 \node [below of = c31, xshift=10pt] (c311) {XHSTT-2014};
 \end{scope}

  \draw[->] (c11.200) |- (c111);
  \draw[->] (c12.200) |- (c121);
  \draw[->] (c12.200) |- (c122);
  \draw[->] (c13.215) |- (c131);
  \draw[->] (c21.200) |- (c211);
  \draw[->] (c22.225) |- (c221);
  \draw[->] (c31.200) |- (c311);

\end{tikzpicture}
    \caption{Educational timetabling problems, formulations, and benchmarks.}
    \label{fig:taxonomy}
\end{small}
\end{figure}
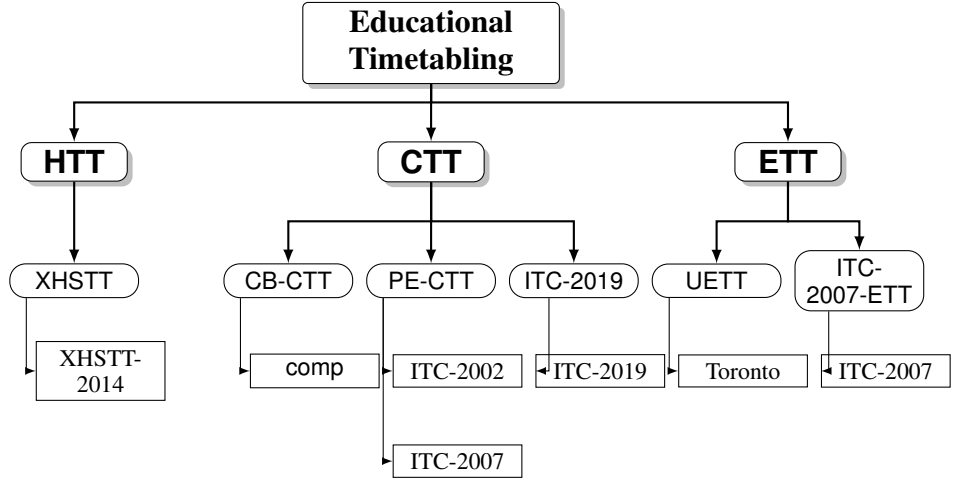

Figure~\ref{fig:taxonomy} shows the taxonomy of the problems presented
in Section~\ref{sec:problems}, and the corresponding formulations
introduced one by one in Section~\ref{sec:formulations-benchmarks}.
For each formulation, the figure reports the datasets used as
benchmarks.

\section{Formulations and Datasets}
\label{sec:formulations-benchmarks}

We introduce the selected formulations and the corresponding datasets
in Sections~\ref{sec:uett}---\ref{sec:itciv}. For each of these
formulations, we present in turn $(i)$ a brief specification, $(ii)$
the benchmarks with their main features, $(iii)$ the file formats and
their usability, $(iv)$ the presence of additional datasets and
instance generators, and $(v)$ some discussion, including the
assessment of the gap w.r.t.\ the complete real-world problem.

From the benchmarks, we identify and remove the instances that are too
easy to be kept in the pool, and their presence results only on a
waste of computational time. We name an instance as easy when all
runs of the top search techniques always find the same score, which is
likely to be the optimal one.

Finally, Section~\ref{sec:other-formulations} is devoted to list and discuss
the other real-world formulations that provide available (and usable) public
datasets.

\subsection{Uncapacitated Examination Timetabling (\uett)}
\label{sec:uett} 

The first formulation that we consider is the classical version of
\ett{} proposed by \citet{CaLL96}, that we name \uett{} (\textsf{U}
for uncapacitated, as explained below). This is a very essential view
of the examination timetabling problem, which extends just slightly the underlying \emph{graph
  coloring} problem, with exams as nodes and periods as colors. 

\paragraph{Short specification}
The main input data of \uett{} is the Boolean-valued \emph{enrollment}
matrix, that stores for each pair $\langle \textrm{student}, \textrm{exam} \rangle$ the
information about whether the student has to take the exam or not.

Two exams with at least one student in common are in conflict, so that
they cannot to be scheduled in the same period. Conflicts are the sole
constraints. In particular, rooms are not taken into account, and for
this reason the problem is known as \emph{uncapacitated}.

The objective function is related to the distance between exams with
students in common. Distances are penalized in the following fixed way: the
cost of scheduling two exams with $k$ students in common at distance of
$1$, $2$, $3$, $4$, and $5$ periods is $16k$, $8k$, $4k$, $2k$, and
$k$, respectively.

\paragraph{Benchmarks}
The \uett{} formulation comes with a dataset of 13 real-world
instances mainly from North American universities, known as Toronto
instances (or Carter's instances), whose main features are illustrated
in Table~\ref{tab:instances} \citep[taken from][]{BCDS21}.

As remarked by \cite{AGVH20}, the instances have some unnecessary data
that could be removed by preprocessing. Indeed, some students are
enrolled in only one exam, so that they can affect neither the
constraints nor the objective. These students are called \emph{noise
  students} by \citeauthor{AGVH20}  In turn, exams taken only by
noise students do not contribute to the constraints and the objective,
and they are called \emph{noise exams}. In Table~\ref{tab:instances},
we report both the total number of students and exams and the
\emph{active} (non-noise) ones.  The two columns \textbf{W} (Workload) represent
the number of exams per active student (average and maximum value).
The rightmost column \textbf{Gd} is the density of the conflict graph, which is
computed as the number of conflicts divided by $n \cdot (n - 1)/2$,
where $n$ is the number of active exams.

\begin{table}[htb]
  \centering
  \begin{footnotesize}
  \begin{tabular}{|l|rr|rr|r|rr|r|}
    \hline    
 Inst.  & \multicolumn{2}{c|}{\textbf{E}} & \multicolumn{2}{c|}{\textbf{S}} & \textbf{P} & \multicolumn{2}{c|}{\textbf{W}} & \textbf{Gd}\\ 
      & total & active &  total & active &          & avg & max & \\ 

        \hline
\textsf{car91}   &     682   &      678  &      16925   &     13516  &   35      &        4.20  &     9	& 0.13 \\
\textsf{car92}   &     543   &      542  &      18419   &     14450  &   32      &        3.84  &     7	& 0.14 \\
\textsf{ear83}   &     190   &      190  &       1125   &      1124  &   24      &        7.21  &     10& 0.27 \\
\textsf{hec92}   &      81   &       81  &       2823   &      2502  &   18      &        4.25  &     7	& 0.42 \\
\textsf{kfu93}   &     461   &      444  &       5349   &      5073  &   20      &        4.92  &     8	& 0.06 \\
\textsf{lse91}   &     381   &      379  &       2726   &      2627  &   18      &        4.16  &     8	& 0.06 \\ 
\textsf{pur93}   &    2627   &      2413 &      30032   &     27405  &   42      &        4.40  &     9 & 0.03 \\
\textsf{rye93}   &     486   &      485  &      11483   &      9458  &   23      &        4.76  &     10& 0.08 \\
\textsf{sta83}   &     139   &      139  &        611   &       611  &   13      &        9.41  &     11& 0.14 \\
\textsf{tre92}   &     261   &      260  &       4360   &      3693  &   23      &        4.03  &     6	& 0.18 \\
\textsf{uta92}   &     622   &      622  &     21266   &     15086  &   35      &        3.91  &     7	& 0.13 \\
\textsf{ute92}   &     184   &      184  &       2750   &      2672  &   10      &        4.41  &     6	& 0.08 \\
\textsf{yor83}   &     181   &      181  &        941   &       940  &   21      &        6.42  &     14& 0.29 \\ 
\hline
  \end{tabular}
  \end{footnotesize}
  \caption{Features of the Toronto benchmark instances. Symbol definition: \textbf{E} (Exams), \textbf{S} (Students), \textbf{P} (Periods), \textbf{W} (Workload), \textbf{Gd} (Graph density).} 
   \label{tab:instances}
\end{table}

\paragraph{File formats and repositories}

Instances are available in plain text and split in two separate files: one
containing the exams and one with the student enrollments. The files
were originally posted via FTP in the website of the University of
Toronto (not active anymore), and are now available at
\url{http://www.cs.nott.ac.uk/~pszrq/data.htm}. The same
instances are posted on \opthub{} with a slightly
modified (more robust), single-file format.

The original data format is unfortunately very fragile, as for example
the accidental insertion of a newline character would result in a
different (but still valid) instance. This has actually happened as discussed below.

\paragraph{Other datasets and generators}

Other instances of \uett{} are available. First, there is a set of 9
instances, called \emph{apocryphal} by \citet{BCDS21}, that are
variants of some of Toronto ones that were created by accidental
perturbation of the original files and used unwittingly in a few
experimental analyses \citep[see][for a discussion about
them]{QBMML09}. Even though they have been considered by a few authors, given that 
they are just arbitrary perturbations of the real
instances, we do not consider them as benchmarks.

Another set of 20 instances, obtained by translating real-world
instances for other examination timetabling formulations, have been made
available by \citeauthor{BCDS21} on \opthub.

Finally, \citet{BCDS21} developed a parametric generator that creates
artificial instances with the prescribed values of the main
features.  A set of 100 generated instances, selected based on feasibility and computational
hardness, are also available on \opthub.

\paragraph{Discussion} 

\uett{} is surely a simplified formulation, as
the authors themselves admit that ``all side constraints have been
removed'' \citep{CaLL96}. Indeed, they list in their original work a set of
constraints that apply to some of the real-world cases, but have been
neglected in the proposed formulation, in order to have a common ground for
many different cases.

Despite its extreme simplicity, or perhaps actually due
to it, \uett{} has been and still is an active subject of studies
(see Section~\ref{sec:res-uett}).  The main reason could also be that the
benchmarks proposed are very challenging. In fact, to the best of our
knowledge, none of such instances has been solved to proven optimality
so far.

\subsection{Post-Enrolment Course Timetabling (\pectt)}
\label{sec:pe-ctt}

The second formulation that we consider is the so-called
Post-Enrolment Course Timetabling (\pectt) problem that is the first
standard formulation of \ctt. It has been proposed within the Metaheuristics 
Network project (2000-04), then used as the subject of
ITC-2002, and used again for ITC-2007 with a slightly more
complex formulation, which is the one discussed here. The full
specification can be found in the work by \citet[\S{3}]{LePM07}.

\paragraph{Short specification}

In \pectt{} it is given a set of events, a set of periods, and a set
of rooms. It is also defined a set of days, such that each period is a
timeslot belonging to one day. Students enroll in events causing
conflicts between them.

Furthermore, there is a set of room features that may be required
by events. Room features and capacity (in terms of seats) together result in a
compatibility relation between rooms and events.

In addition, it is defined a precedence relation between events, such
that some events must be scheduled before others. Finally, the last
constraints are the ones originated from an unavailability relation,
stating that an event cannot be scheduled in some specified periods.

The objective function is composed by three components that penalize
the following cases: $(i)$ a student attending an event in
the last timeslot of a day, $(ii)$ a student
attending three (or more) events in successive timeslots in the
same day, $(iii)$ a student attending only one event
in a day.

\paragraph{Benchmarks}

There are two datasets that can be considered as consolidated
benchmarks for \pectt, which are the ones coming from the competitions
ITC-2002 and ITC-2007. The dataset from ITC-2002 is on a simplified
version of the problem that does not consider precedences and
unavailabilities.

\begin{table}[tbh]
  \centering
\begin{footnotesize}
    \begin{tabular}{|l|rrrrrrrr|}\hline
            Inst. & \textbf{E} & \textbf{R} & \textbf{S} & \textbf{Ro} & \textbf{Cgd} & \textbf{SE} & \textbf{ES} & \textbf{RE} \\ \hline
      \textsf{01} & 400 & 10 & 200 & 0.89 & 0.20 & 8.88 & 17.75 & 1.96 \\
      \textsf{02} & 400 & 10 & 200 & 0.89 & 0.21 & 8.61 & 17.23 & 1.92 \\
      \textsf{03} & 400 & 10 & 200 & 0.89 & 0.23 & 8.85 & 17.70 & 3.42 \\
      \textsf{04} & 400 & 10 & 300 & 0.89 & 0.23 & 13.07 & 17.43 & 2.45 \\
      \textsf{05} & 350 & 10 & 300 & 0.78 & 0.31 & 15.24 & 17.78 & 1.78 \\
      \textsf{06} & 350 & 10 & 300 & 0.78 & 0.26 & 15.23 & 17.77 & 3.59 \\
      \textsf{07} & 350 & 10 & 350 & 0.78 & 0.21 & 17.48 & 17.48 & 2.87 \\
      \textsf{08} & 400 & 10 & 250 & 0.89 & 0.17 & 10.99 & 17.58 & 2.93 \\
      \textsf{09} & 440 & 11 & 220 & 0.89 & 0.17 & 8.68 & 17.36 & 2.58 \\
      \textsf{10} & 400 & 10 & 200 & 0.89 & 0.20 & 8.89 & 17.78 & 3.49 \\
      \textsf{11} & 400 & 10 & 220 & 0.89 & 0.20 & 9.58 & 17.41 & 2.07 \\
      \textsf{12} & 400 & 10 & 200 & 0.89 & 0.20 & 8.79 & 17.58 & 1.96 \\
      \textsf{13} & 400 & 10 & 250 & 0.89 & 0.21 & 11.06 & 17.69 & 2.43 \\
      \textsf{14} & 350 & 10 & 350 & 0.78 & 0.25 & 17.42 & 17.42 & 3.08 \\
      \textsf{15} & 350 & 10 & 300 & 0.78 & 0.25 & 15.07 & 17.58 & 2.19 \\
      \textsf{16} & 440 & 11 & 220 & 0.89 & 0.18 & 8.88 & 17.75 & 3.17 \\
      \textsf{17} & 350 & 10 & 300 & 0.78 & 0.31 & 15.15 & 17.67 & 1.11 \\
      \textsf{18} & 400 & 10 & 200 & 0.89 & 0.21 & 8.78 & 17.56 & 1.75 \\
      \textsf{19} & 400 & 10 & 300 & 0.89 & 0.20 & 13.28 & 17.71 & 3.94 \\
      \textsf{20} & 350 & 10 & 300 & 0.78 & 0.25 & 14.99 & 17.49 & 3.43 \\\hline
    \end{tabular}
  \end{footnotesize}
\caption{Features of the ITC-2002 benchmark instances. Symbol definition: \textbf{E} (Events), \textbf{R}  (Rooms), \textbf{S}  (Students), \textbf{Ro} (Room occupancy), \textbf{Cgd} (Conflict graph density), \textbf{SE} (average Students per Exam), \textbf{ES} (average Exam per Student), \textbf{RE}  (average suitable Rooms per Event).}
\label{tab:pectt-instances1}
\end{table}

\begin{table}
  \centering
  \begin{footnotesize}
\begin{tabular}{|l|rrrrrrrrrr|} \hline
Inst. & \textbf{E} & \textbf{R} & \textbf{S} & \textbf{Ro} & \textbf{Cgd} & \textbf{SE} & \textbf{ES} & \textbf{RE} & \textbf{TE} & \textbf{P}\\ \hline
\textsf{01} & 400 & 10 & 500 & 0.89 & 0.34 & 26.27 & 21.02 & 4.08 & 0.56 & 0.10 \\
\textsf{02} & 400 & 10 & 500 & 0.89 & 0.37 & 26.29 & 21.03 & 3.95 & 0.57 & 0.09 \\
\textsf{03} & 200 & 20 & 1000 & 0.22 & 0.47 & 66.92 & 13.38 & 5.05 & 0.56 & 0.10 \\
\textsf{04} & 200 & 20 & 1000 & 0.22 & 0.52 & 66.98 & 13.40 & 6.40 & 0.57 & 0.10 \\
\textsf{05} & 400 & 20 & 300 & 0.44 & 0.31 & 15.69 & 20.92 & 6.80 & 0.56 & 0.37 \\
\textsf{06} & 400 & 20 & 300 & 0.44 & 0.30 & 15.55 & 20.73 & 5.07 & 0.56 & 0.35 \\
\textsf{07} & 200 & 20 & 500 & 0.22 & 0.53 & 33.67 & 13.47 & 1.58 & 0.39 & 0.10 \\
\textsf{08} & 200 & 20 & 500 & 0.22 & 0.52 & 34.58 & 13.83 & 1.91 & 0.38 & 0.10 \\
\textsf{09} & 400 & 10 & 500 & 0.89 & 0.34 & 26.78 & 21.43 & 2.91 & 0.56 & 0.11 \\
\textsf{10} & 400 & 10 & 500 & 0.89 & 0.38 & 26.23 & 20.98 & 3.20 & 0.56 & 0.10 \\
\textsf{11} & 200 & 10 & 1000 & 0.44 & 0.50 & 68.04 & 13.61 & 3.38 & 0.56 & 0.10 \\
\textsf{12} & 200 & 10 & 1000 & 0.44 & 0.58 & 68.04 & 13.61 & 3.36 & 0.57 & 0.10 \\
\textsf{13} & 400 & 20 & 300 & 0.44 & 0.32 & 15.90 & 21.19 & 8.68 & 0.56 & 0.34 \\
\textsf{14} & 400 & 20 & 300 & 0.44 & 0.32 & 15.64 & 20.86 & 7.56 & 0.56 & 0.36 \\
\textsf{15} & 200 & 10 & 500 & 0.44 & 0.54 & 32.63 & 13.05 & 2.23 & 0.38 & 0.10 \\
\textsf{16} & 200 & 10 & 500 & 0.44 & 0.46 & 34.10 & 13.64 & 1.74 & 0.39 & 0.11 \\
\textsf{17} & 100 & 10 & 500 & 0.22 & 0.71 & 97.67 & 19.53 & 2.77 & 0.57 & 0.12 \\
\textsf{18} & 200 & 10 & 500 & 0.44 & 0.65 & 51.42 & 20.57 & 3.47 & 0.57 & 0.10 \\
\textsf{19} & 300 & 10 & 1000 & 0.67 & 0.47 & 44.78 & 13.44 & 3.66 & 0.56 & 0.10 \\
\textsf{20} & 400 & 10 & 1000 & 0.89 & 0.28 & 33.92 & 13.57 & 3.73 & 0.56 & 0.10 \\
\textsf{21} & 500 & 20 & 300 & 0.56 & 0.23 & 12.40 & 20.67 & 7.36 & 0.57 & 0.36 \\
\textsf{22} & 600 & 20 & 500 & 0.67 & 0.26 & 17.42 & 20.90 & 5.65 & 0.56 & 0.39 \\
\textsf{23} & 400 & 20 & 1000 & 0.44 & 0.44 & 53.42 & 21.37 & 2.89 & 0.78 & 0.12 \\
\textsf{24} & 400 & 20 & 1000 & 0.44 & 0.31 & 33.34 & 13.34 & 1.59 & 0.55 & 0.72 \\\hline
\end{tabular}
\end{footnotesize}
\caption{Features of ITC-2007 benchmark instances.  Symbol definition: \textbf{E} (Events), \textbf{R}  (Rooms), \textbf{S}  (Students), \textbf{Ro} (Room occupancy), \textbf{Cgd} (Conflict graph density), \textbf{SE} (average Students per Exam), \textbf{ES} (average Exam per Student), \textbf{RE}  (average suitable Rooms per Event), \textbf{TE} (average availability of Timeslots for Event), \textbf{P} (average ratio of Precedences per Event).}
\label{tab:pectt-instances2}
\end{table}

The main features of the instances are illustrated in
Tables~\ref{tab:pectt-instances1} and~\ref{tab:pectt-instances2}. The
number of periods is 45 for all instances, thus it is not listed in
the tables.  All instances are artificial and obtained by a generator.
In addition, they have the peculiarity of having been generated in such a way
that at least one \emph{perfect} (zero cost) solution exists.

For ITC-2007 instances it is generally more difficult than ITC-2002 ones to
find a feasible solution. Indeed, looking at the tables, we see that
they have a higher density (column \textbf{Cgd}), in addition to unavailabilities
and precedences that are absent in the ITC-2002 version.

\paragraph{File formats and repositories}

All instances are available in a lengthy text-only format, in which
all elements of the matrices are written explicitly, one per line. As
a consequence, the files are easy to parse, but rather verbose, not
human readable, and fragile. They are available at the websites of the
competitions, reachable from the PATAT conference website
(\url{http://patatconference.org/}). Instances are available also from
\opthub.

\paragraph{Other datasets and generators}

Two other datasets are publicly available, and they have been
considered in some papers, though less frequently than the two
mentioned above. The first one is a dataset proposed by the
Metaheuristics Network \citep[see][]{RSBC03} before the competitions,
using the simplified formulation of ITC-2002, which are available at
\url{http://iridia.ulb.ac.be/supp/IridiaSupp2002-001}. For recent
results on these instances see for example \citep{GoKS17}.

The second dataset, including much larger instances, has been
introduced by \citet{LePa07} with the aim of having more difficult
cases and is available at \url{http://www.rhydlewis.eu/hardTT/}.
Indeed, for these instances, feasibility is quite difficult to be
obtained, and the comparison is on the number of violations rather
than on the objective function. For results on these instances see for
example \citep{CeDS12}.

The generator used for the instances of ITC-2002 and ITC-2007 was
never made public, and no other one has been developed and made
available for this formulation.

\paragraph{Discussion} 

Like \uett{} discussed in the previous section, \pectt{} is rather a
simplified formulation with respect to the real-world problem, as many
aspects are deliberately removed in order to make the problem more
manageable \cite[see][\S{5} for a discussion about the neglected
features]{LePM07}.

In addition, as shown in Tables~\ref{tab:pectt-instances1}
and~\ref{tab:pectt-instances2}, the diversity of the features of the
instances is quite limited. For example, all instances have exactly 45
periods divided in 5 days of 9 timeslots each, with no variability at
all. Similarly, the number of rooms is restricted to just three
different values, namely 10, 11, and 20.

\subsection{Curriculum-Based Course Timetabling (\cbctt)}
\label{sec:cb-ctt}

The next formulation that we discuss is the so-called Curriculum-based
Course Timetabling (\cbctt), which has been proposed by \cite{DiSc03},
and subsequently adopted, in a slightly modified version, as the third
track of ITC-2007.

The name of the problem comes from the fact that conflicts are
determined by predefined curricula, opposed to the use of explicit
student enrolments as in \pectt. This is however not the most
important difference between \pectt{} and \cbctt{} as the notions of
student and curriculum are formally interchangeable, because a student
can be expressed as a curriculum and vice versa. On the contrary, the
main difference stems from the notion of \emph{course} as a set of
lectures that is absent in \pectt. Many constraints and
objectives in \cbctt{} are defined at the level of a course, whereas in
\pectt{} constraints and objectives are always expressed at the level
of the single event/lecture.

A few variants of this formulation have been subsequently proposed
by \citet{BDDS12}. The most studied one however remains the one used
for ITC-2007 \citep[named UD2 in][]{BDDS12}, which is thus the one
that we consider here. The full description is provided by
\citet{DiMS07}.

\paragraph{Short specification}

As mentioned above, the key notions of \cbctt{} are courses and
curricula.  Each course consists of a fixed number of \emph{lectures}
to be scheduled in different periods. A course is attended by a number
of \emph{students}, and is taught by a \emph{teacher}. For each
course, there are a minimum number of days over which the lectures of
the course should be spread.  Moreover, there are some unavailable
periods in which the course cannot be scheduled.

Like in \pectt{}, we are given a number of periods divided in days and
timeslots in the day.  Each \emph{room} has a \emph{capacity},
specified as the number of available seats, but no other features.

A \emph{curriculum} is a group of courses that potentially have
students in common. As a consequence, lectures of courses belonging to
the same curriculum are in conflict and cannot be scheduled in
the same period. Two courses are in conflict also if they are taught
by the same teacher.

The hard constraints regard conflicts, teacher availability and room
occupancy.  The objective function (soft constraints) is composed by
four components that penalize the following cases: $(i)$ the capacity
of the room assigned to a lecture is less than the number of students
attending the course, $(ii)$ the lectures of a course are not spread
into the given minimum number of days, $(iii)$ a lecture is isolated,
i.e., not adjacent to any other in the same curriculum, $(iv)$ the
lectures of a course are not given all in the same room.

\paragraph{Benchmarks}

Quite a few real-world datasets are available for the formulation. By
far the most used one is the one from ITC-2007, known as the
\texttt{comp} dataset that we consider as benchmark.

Table~\ref{tab:cbctt-instances} \citep[taken from][]{BDDS12} shows the
main features of these instances. It could be noticed that we removed
instances \textsf{comp11} and \textsf{comp15}. They have been remove
for different reasons: \textsf{comp11} is an easy one, as all
competitive search methods always find a solution of cost zero, while
\texttt{comp15} is actually identical to \texttt{comp03} in the
problem variant UD2 that we consider here.

\begin{table}
\centering
\begin{footnotesize}
\begin{tabular}{|l|ccccccccccc|}\hline
Inst.~~&~\textbf{C}~&~\textbf{L}~&~\textbf{R}~&~\textbf{PpD}~&~\textbf{D}~&~\textbf{Cu}~&~\textbf{MML}~&~\textbf{Co}~&~\textbf{TA}~&~\textbf{CL}~&~\textbf{RO}~\\ \hline
\textsf{comp01} & 30 & 160 & 6 & 6 & 5 & 14 & 2-5 & 13.2 & 93.1 & 3.24 & 88.9\\
\textsf{comp02} & 82 & 283 & 16 & 5 & 5 & 70 & 2-4 & 7.97 & 76.9 & 2.62 & 70.8\\
\textsf{comp03} & 72 & 251 & 16 & 5 & 5 & 68 & 2-4 & 8.17 & 78.4 & 2.36 & 62.8\\
\textsf{comp04} & 79 & 286 & 18 & 5 & 5 & 57 & 2-4 & 5.42 & 81.9 & 2.05 & 63.6\\
\textsf{comp05} & 54 & 152 & 9 & 6 & 6 & 139 & 2-4 & 21.7 & 59.6 & 1.8 & 46.9\\
\textsf{comp06} & 108 & 361 & 18 & 5 & 5 & 70 & 2-4 & 5.24 & 78.3 & 2.42 & 80.2\\
\textsf{comp07} & 131 & 434 & 20 & 5 & 5 & 77 & 2-4 & 4.48 & 80.8 & 2.51 & 86.8\\
\textsf{comp08} & 86 & 324 & 18 & 5 & 5 & 61 & 2-4 & 4.52 & 81.7 & 2 & 72\\
\textsf{comp09} & 76 & 279 & 18 & 5 & 5 & 75 & 2-4 & 6.64 & 81 & 2.11 & 62\\
\textsf{comp10} & 115 & 370 & 18 & 5 & 5 & 67 & 2-4 & 5.3 & 77.4 & 2.54 & 82.2\\
\textsf{comp12} & 88 & 218 & 11 & 6 & 6 & 150 & 2-4 & 13.9 & 57 & 1.74 & 55.1\\
\textsf{comp13} & 82 & 308 & 19 & 5 & 5 & 66 & 2-3 & 5.16 & 79.6 & 2.01 & 64.8\\
\textsf{comp14} & 85 & 275 & 17 & 5 & 5 & 60 & 2-4 & 6.87 & 75 & 2.34 & 64.7\\
\textsf{comp16} & 108 & 366 & 20 & 5 & 5 & 71 & 2-4 & 5.12 & 81.5 & 2.39 & 73.2\\
\textsf{comp17} & 99 & 339 & 17 & 5 & 5 & 70 & 2-4 & 5.49 & 79.2 & 2.33 & 79.8\\
\textsf{comp18} & 47 & 138 & 9 & 6 & 6 & 52 & 2-3 & 13.3 & 64.6 & 1.53 & 42.6\\
\textsf{comp19} & 74 & 277 & 16 & 5 & 5 & 66 & 2-4 & 7.45 & 76.4 & 2.42 & 69.2\\
\textsf{comp20} & 121 & 390 & 19 & 5 & 5 & 78 & 2-4 & 5.06 & 78.7 & 2.5 & 82.1\\
\textsf{comp21} & 94 & 327 & 18 & 5 & 5 & 78 & 2-4 & 6.09 & 82.4 & 2.25 & 72.7\\ \hline
\end{tabular}
\end{footnotesize}
\caption{Features of comp benchmark instances. Symbol definition: 
\textbf{C} (Courses), \textbf{L} (total Lectures), \textbf{R} (Rooms), \textbf{PpD} (Periods per Day), \textbf{D} (Days), 
\textbf{Cu} (Curricula), \textbf{MML} (Min and Max
Lectures per day per curriculum), \textbf{Co} (average number of Conflicts), 
\textbf{TA} (average Teacher Availability), \textbf{CL} (average number of Lectures
per Curriculum per day), \textbf{RO} (average Room Occupation).}
\label{tab:cbctt-instances}
\end{table}

\paragraph{File formats and repositories}
Instances are available in an ad-hoc text-only format, which is
reasonably human-readable. There are actually two versions of the
format, the original \texttt{.ctt} one used for the competition, and
the newer \texttt{.ectt} (\texttt{e} for extended) proposed by
\cite{BDDS12} that includes additional data necessary for the other
versions of the problem.

Instances are available on \opthub{} in \texttt{.ectt} format along
with several results from the literature. The solutions were imported
from the original \cbctt{} website (\url{satt.diegm.uniud.it/ctt}) not
available anymore.

\paragraph{Other datasets and generators}

A few other datasets of real-world instances coming mainly from Italian universities are available on \opthub{}.

An instance generator has been developed by \citet{BMPR08}, which has
been subsequently revised by \citet{LoSc10,LoSc13} in such a way to obtain more
realistic instance, in particular more similar to the \texttt{comp}
dataset. The latter generator has been further refined by
\citet{DMSSS21} in order to enlarge the region of the instance space
covered by the generated instances. The generator by \citeauthor{DMSSS21} is available at
\url{https://cdlab-artis.dbai.tuwien.ac.at/papers/cb-ctt/}.

\paragraph{Discussion} 

Like \uett{} and \pectt{}, the \cbctt{} formulation is a judicious
simplification of the original problem. Constraints and
objectives included in the formulation have been selected among the
long list of real ones to be general and simple enough, but also
representative of the various types of restrictions. For example, the
objective on room stability for the lectures of a course is not
particularly important in practice, but it represents a set of
limitations that involve the use of rooms in different periods.
Without this objective the management of the rooms could have been
done independently for each period, which would have resulted in an
oversimplification of the problem.

The \texttt{comp} instances are extracted from various departments of
University of Udine (Italy), so that the values of the main features
are quite diverse. The additional instances come also from different
universities, so that they are yet broader in size and structure.

\subsection{Examination Timetabling (\itcii)}
\label{sec:itcii}

The next formulation that we include in our study is the \ett{}
proposed for ITC-2007 (Track 1). This formulation, even though it does
not consider all practical features, is much more realistic than the
uncapacitated version discussed in Section~\ref{sec:uett}. Indeed, it
includes several novel features collected from the activity of a
commercial software in use in many British universities. The full
specification is provided by \citet{MMBPQ07}.

\paragraph{Short specification}

Like other formulations, the time horizon is divided in a number of
periods, each one belonging to a day. The novelty is that periods have
a specific length (in minutes) and can have a penalty for scheduling
exams in it.

As usual, rooms have a capacity and might be undesired, in the 
sense that there is a penalty for their use, like periods.

For each exam, a length of execution is given, so that it is
compatible only with periods of sufficient duration. In addition, an
exam might require to be scheduled in a dedicated room, otherwise it
can share the room with other exams. For each exam, it is also given the set of
students enrolled for it.

For some pairs of exams a precedence rule is specified, stating that
one exam must be scheduled after, at the same time, or at a different
time with respect to the other one.

The objective function is composed by the following components (soft
constraints): $(i)$ a student taking two exams in consecutive periods
in the same day, $(ii)$ a student taking two exams in the same day
$(iii)$ a student taking two exams within a fixed number of periods
(spread), $(iv)$ exams in the same room with mixed durations, $(v)$ an
exam with many students scheduled towards the end of the planning
horizon, $(vi)$ an exam scheduled in an undesired period or an
undesired room.

The weights of the soft constraints vary from case to case, and
are included in the input file of each instance.

\paragraph{Benchmarks}

A dataset composed of 12 instances from British universities was
released for ITC-2007. The main features of these
instances\footnote{Note that the number of exams reported by
  \citet[Table 2]{BuBy16} is overestimated as they consider the
  largest student identifier, but some numbers are missing in the
  file. } are summarized in Table~\ref{tab:examtt-features}
\citep[adapted from][]{BaSU17}.

\begin{table}
\centering
\begin{footnotesize}
    \begin{tabular}{|c|r|r|r|r|r|r|r@{}c@{}l|r|r|r|r|r|r|}
        \hline
        Inst.  &   \textbf{E} & \textbf{S} & \textbf{P} & \textbf{R} &\textbf{P$_{HC}$} & \textbf{R$_{HC}$} &  \textbf{FP} & / & \textbf{FE} & \textbf{PS} & \textbf{Cd} & \textbf{ER} & \textbf{SE} & \textbf{S/Cap} & \textbf{PC}   \\       \hline
\textsf{1} & 607  & 7883  & 54 & 7  & 12 & 0  & 30 &/&  100     & 5        & 0.05 & 1.61  & 53.3 & 0.75 & 15.9 \\
\textsf{2} & 870  & 12484 & 40 & 49 & 8  & 2  & 30 &/&  250     & 1        & 0.01 & 0.44  & 43.0 & 0.23 & 26.5 \\
\textsf{3} & 934  & 16365 & 36 & 48 & 82 & 15 & 20 &/& 200      & 4        & 0.03 & 0.54  & 65.5 & 0.33 & 34.1 \\
\textsf{4} & 273  & 4421  & 21 & 1  & 20 & 0  & 10 &/& 50       & 2        & 0.15 & 13.00 & 79.6 & 0.86 & 19.1 \\
\textsf{5} & 1018 & 8719  & 42 & 3  & 27 & 0  & 30 &/& 250      & 5        & 0.01 & 8.08  & 33.6 & 0.34 & 72.6 \\
\textsf{6} & 242  & 7909  & 16 & 8  & 22 & 0  & \textbf{30} &/& 25   & \textbf{20} & 0.06 & 1.9  & 76.31 & 0.56 & 35.0 \\
\textsf{7} & 1096 & 13795 & 80 & 15 & 28 & 0  & 30 &/& 250      & 10       & 0.02 & 0.91  & 41.5 & 0.22 & 43.6 \\
\textsf{8} & 598  & 7718  & 80 & 8  & 20 & 1  & \textbf{100} &/& 250 & 15       & 0.05 & 0.93  & 52.5 & 0.43 & 177.0\\ 
\textsf{9} & 169  & 624   & 25 & 3  & 10 & 0  & 10  &/& 100     & 5        & 0.08 & 2.25  & 15.0 & 0.60 & 32.8 \\
\textsf{10} & 214 & 1415  & 32 & 48 & 58 & 0  & 10 &/& 100      & 20       & 0.05 & 0.14  & 36.7 & 0.13 & 89.4 \\
\textsf{11} & 934 & 16365 & 26 & 40 & 83 & 15 & 20 &/& 400      & 4        & 0.03 & 0.90  & 65.5 & 0.48 & 51.1 \\
\textsf{12} & 78  & 1653  & 12 & 50 & 9  & 7  & 5  &/& 25       & 5        & 0.18 & 0.13  & 47.2 & 0.20 & 36.7 \\ \hline
    \end{tabular}
 \end{footnotesize}
    \caption{Features of the ITC-2007 benchmark instances. Symbol definition:
\textbf{E} (Exams), \textbf{S} (Students), \textbf{P} (Periods), \textbf{R} (Rooms), \textbf{P$_{HC}$} (Periods Hard Constraints), \textbf{R$_{HC}$}  (Rooms Hard Constraints),  \textbf{FP} / \textbf{FE} (Frontload Periods / Frontload Exams), \textbf{PS} (Periods Spread), \textbf{Cd} (Conflict density), \textbf{ER} (Exams per Room), \textbf{SE} (Students per Exam), \textbf{S/Cap} (Students to Capacity Ratio), \textbf{PC} (Period Conflict). The values highlighted in boldface are incoherent with the number of available periods.
\label{tab:examtt-features}}
\end{table}

\paragraph{File formats and repositories}

The file format is a single-file text-only one, created ad-hoc for the
ITC-2007 competition. Although the format is better engineered than
the simple one of previous competitions, still there are some
fragilities, as it is witnessed by the presence of incoherent data in
two of the competition instances\footnote{Instances 6 and 8 have a
  number of Front Period (both instances) and Period Spread (instance
  6 only) larger than the number of periods (highlighted in boldface).
  This implies that the corresponding soft constraints are always
  violated, independently of the solution.}.  Fortunately, these
inconsistencies do not affect the significance of those instances but
the meaning of the two soft constraint types involved is irrelevant
since they are unsatisfiable.

The files are available from the ITC-2007 website and also from
\opthub, where there are also a few solutions listed.

\paragraph{Other datasets and generators}

A set of 8 instance coming from Yeditepe University and proposed by
\cite{OzEr05} have been translated from their original format to the
ITC-2007 one by \cite{PaOz10}. They are available at
\url{http://www.cs.nott.ac.uk/~pszajp/timetabling/exam/}. 

These
instance are relatively small, with a maximum size of 210 exams. In
addition, they are obtained from a simpler formulation, so that many
features of the \itcii{} formulation are unused.  Up to our knowledge,
no other real-world instances have been introduced later on for the problem.

An instance generator has been developed by \citet{BaSU17} for tuning
purposes, and a set of 50 challenging artificial instance has been
made public on \opthub.

The original solution checker provided for ITC-2007 is not available anymore,
but solutions can be validated from \opthub.

\paragraph{Discussion}

As mentioned by \citet{MMBPQ07}, this formulation is a significant
step forward the use of complete formulations for standard problems.
Indeed, with respect to its predecessors it includes many novel real-world features, in particular of
British universities, even though, for the aim of simplicity, some aspects
are still left out.

This is also the first formulation of our list that has the weights of
the different objectives written in the instance, rather than fixed for
all scenarios. As written by \citet{MMBPQ07}: ``this is motivated by our
experience that different institutions do indeed have different
weights, and so no one set would be completely useful''. Still from
\citet{MMBPQ07}: ``We hope that this will encourage the development of
solvers that are robust rather than potentially over-tuned to one
particular set of weights for a dataset.''

\subsection{High-School Timetabling (\xhstt)}
\label{sec:xhstt}

Our next formulation was introduced by \cite{PADK+12} as an attempt to
create a unified formulation and data format for the \htt{} problem.
The proposed formulation, called \xhstt{}, is extremely rich and the
intent is to avoid, differently from the previous formulations, any
concession to judicious simplifications.  In fact, the proposing team
is composed by researchers from various countries, with the aim of
including the features coming from as many different situations as
possible all around the world.

\xhstt{} has also been used as the subject of the ITC-2011
competition, which has led to a boost for its spread in the scientific
community. In fact, \xhstt{} is the most popular formulation compared
to other ones among the community, and it has drawn the attention of
many researchers, in particular after ITC-2011.  In addition, the
dataset is diverse and quite challenging, also compared to previous ones.

Over the years, several versions of the archive have been collected in
the XHSTT project, each one mainly based on the previous one with some
improvements on the current instances (name change, format
simplification, error correction, redundancy removal, \dots) and some
new instances. As a consequence, in some cases authors have competed
on slightly different versions of the same instances, so that a
comprehensive and fair comparison has been made not possible.  Indeed,
\cite{KrSS15} wrote: ``such updates to the format make it hard for
researchers to compare computational results with those previously
reported''.

For the full problem specification, we refer to the work by \citet{PADK+12}
and to the \xhstt{} website \url{https://www.utwente.nl/hstt/}, which contains
also some updates with respect to the original formulation.

\paragraph{Short specification}

As mentioned above, the formulation is complex, so that it is quite
difficult to discuss it in brief. Basically, it includes three
types of entities: times, resources (students, classes, teachers, 
and rooms), and events. For each of these three, it is possible to 
define sets of atomic elements, and use these sets to express complex 
constraints and objectives.

In \xhstt{} there are 15 different types of constraints, which range from 
spreading lectures in the week, to student idle times, to preferences and 
unavailabilities. For brevity, we do not list them here and refer again 
to the \xhstt{} website for their comprehensive specification. 
Each individual constraint can be declared either hard (Required) or 
soft (non-Required). 

\paragraph{Benchmarks}

As benchmarks we consider the current version of the archive at the
\xhstt{} website, called XHSTT-2014. As mentioned in the website:
``XHSTT-2014 contains a carefully selected subset of the instances
collected during this project, in their most up-to-date form''.

\begin{table}[htbp]
  \centering
  \begin{footnotesize}
    \begin{tabular}{|l|r|r|r|r|r|r|r|} \hline
    Inst. & \textbf{T} &\textbf{Te} & \textbf{R} &\textbf{C} & \textbf{S} & \textbf{E} & \textbf{D} \\\hline
    \textsf{AU-BG-98} & 40    & 56    & 45    & 30    &   ---    & 387   & 1564 \\
    \textsf{AU-SA-96} & 60    & 43    & 36    & 20    &   ---    & 296   & 1876 \\
    \textsf{AU-TE-99} & 30    & 37    & 26    & 13    &    ---   & 308   & 806 \\
    \textsf{BR-SA-00} & 25    & 14    &    ---   & 6     &   ---    & 63    & 150 \\
    \textsf{BR-SM-00} & 25    & 23    &  ---     & 12    &    ---   & 127   & 300 \\
    \textsf{BR-SN-00} & 25    & 30    &   ---    & 14    &    ---   & 140   & 350 \\
    \textsf{DK-FG-12} & 50    & 90    & 69    &   ---    & 279   & 1077  & 1077 \\
    \textsf{DK-HG-12} & 50    & 100   & 71    &  ---     & 523   & 1235  & 1235 \\
    \textsf{DK-VG-09} & 60    & 46    & 53    &  ---     & 163   & 918   & 918 \\
    \textsf{ES-SS-08} & 35    & 66    & 4     & 21    &   ---    & 225   & 439 \\
    \textsf{FI-PB-98} & 40    & 46    & 34    & 31    &    ---   & 387   & 854 \\
    \textsf{FI-WP-06} & 35    & 18    & 13    & 10    &  ---     & 172   & 297 \\
    \textsf{FI-MP-06} & 35    & 25    & 25    & 14    &   ---    & 280   & 306 \\
    \textsf{GR-PA-08} & 35    & 19    &   ---    & 12    &    ---   & 262   & 262 \\
    \textsf{IT-I4-96} & 36    & 61    &  ---     & 38    &  ---     & 748   & 1101 \\
    \textsf{KS-PR-11} & 62    & 101   &    ---   & 63    &    ---   & 809   & 1912 \\
    \textsf{NL-KP-03} & 38    & 75    & 41    & 18    & 453   & 1156  & 1203 \\
    \textsf{NL-KP-05} & 37    & 78    & 42    & 26    & 498   & 1235  & 1272 \\
    \textsf{NL-KP-09} & 38    & 93    & 53    & 48    &  ---     & 1148  & 1274 \\
    \textsf{UK-SP-06} & 25    & 68    & 67    & 67    &     ---  & 1227  & 1227 \\
    \textsf{US-WS-09} & 100   & 134   & 108   &   ---    &    ---   & 628   & 6354 \\
    \textsf{ZA-LW-09} & 148   & 19    & 2     & 16    &      --- & 185   & 838 \\
    \textsf{ZA-WD-09} & 42    & 40    &     ---  & 30    &    ---   & 278   & 1353 \\\hline
    \end{tabular}
 \end{footnotesize}
 \caption{Features of the \xhstt{} benchmark instances. Symbol definition:
\textbf{T} (Times),  \textbf{Te} (Teachers), \textbf{R} (Rooms), \textbf{C} (Classes), \textbf{S} (Students), \textbf{E} (Events), \textbf{D} (Duration).}
  \label{tab:xhstt-features}
\end{table}

This archive is composed of 25 instances. We removed two of them,
namely GR-H1-97 and GR-P3-10, for which a perfect solution (i.e., having zero cost) 
can be easily
obtained. The main features of the remaining 23 instances are shown in
Table~\ref{tab:xhstt-features} taken from the \xhstt{} website. The
symbol ``---'' means that the corresponding resource group is omitted
in the particular instance.

\paragraph{File formats and repositories}

Instances are written in an XML file format, which includes also a
metadata part. Thanks to the flexibility of XML, many instances and
solutions can be inserted in a single file.

All instances, lower bounds, and best solutions are available at the
\xhstt{} website, including a checker that validates a solution and
writes a report of the corresponding violations.

\paragraph{Other datasets and generators}

Other instances have been contributed from the community over the
years and they have been included in the \xhstt{} website, but they
are currently considered less interesting. There are also a few
artificial ones obtained by translating instances from other
formulations, which do not use most of the constraint types. All the
instances are available from the \xhstt{} website. Up to our
knowledge, no artificial instance generator is available.

\paragraph{Discussion}

As mentioned above, \xhstt{} is a full-fledged real-world problem with
all possible constraints and objectives included. The spirit of this
effort is to consider all possible constraints in use somewhere in the
globe, allowing the possibility to produce instances that use only a
subset of the constraints for its specification. The formulation is
still evolving, with student sectioning and different campuses as
candidate new features.

A drawback of this choice is that is it rather labor demanding to
implement an effective solver for the complete specification of
\xhstt{}.  However, a solver could also be developed to deal with only
a subset of the possible constraint types.

A limit of the benchmark dataset is that nowadays many instances are
solved to proven optimality, so that the competition is moved mainly
to the performance of solvers under specific timeouts. An alternative
standard formulation, which has recently gained some attention, is the
Brazilian one introduced by \cite{SaCo17}, mentioned in
Section~\ref{sec:other-formulations}, and described in the survey by
\citet{TGKS21}.

\subsection{University Course Timetabling (\itciv)}
\label{sec:itciv}

Our last standard formulation is the one of the \ctt{} problem
proposed by \citet{MuRM18} for the ITC-2019 competition, that we call
\itciv. This formulation actually represents a combination of \ctt{}
with the student sectioning problem. The formulation is indeed rather
rich and structured, and it represents a big step forward bridging the
gap between theory and practice of timetabling research. Nonetheless,
it still cannot be considered a totally complete problem, as the
authors themselves write ``to make the problems more attractive, we
remove some of the less important aspects of the real-life data while
retaining the computational complexity of the problems''.

\paragraph{Short specification}

\itciv{} consists in sectioning students into classes based on courses
enrollments, and then assigning classes to available periods and
rooms, respecting various constraints and preferences.

The main novelty is that courses may have a complex structure of
classes, with one or more configurations, further divided in subparts,
and parent-child relationship between classes. For each class, it is
also specified the list of possible periods and rooms for meetings.

The other remarkable feature is that the timetable may differ from
week to week, differently from \cbctt{} and \pectt{} where the very
same weekly timetable is replicated for the whole semester.  This
feature is present in many practical situations as it allows the
institution to gain flexibility in the organization.

Lastly, there are many \emph{distribution} constraints that are
evaluated between individual pairs of classes, or all classes as a
whole. Distribution constraints may affect the time of the day, the
days of the week, the week of the semester, or the room assigned
\citep[see][\S3.5]{MuRM18}.

A penalty is associated to the selection of a room and a period for a
class, so that the objective function is composed by four main
components: $(i)$ class/period penalization, $(ii)$ class/room
penalization, $(iii)$ violations of distribution constraints, and
$(iv)$ student conflicts.

\paragraph{Benchmarks}

Instances come from ten institutions, including Purdue University
(USA), Masaryk University (Czech Republic), AGH University of Science
and Technology (Poland), and Istanbul K\"ultu\"ur University (Turkey).
The real-life data was properly anonymized and simplified
as discussed below.

The dataset is composed by 30 instances from ITC-2019 (10 early, 10
middle, 10 late) with very different features in terms of size of the
problem (number of classes, students and rooms), room utilization,
student course demand, course structure, time patterns, travel times
and distribution constraints. 
Such diversity reflects the different
sources of the data, both for the type of institution
(school/faculty/entire university) and geographical position.
Table~\ref{tab:itc2019-features}  reports a selection of the  instance features 
available from a more comprehensive list published on the competition website.

\begin{table}[htbp]
\centering
\begin{footnotesize}
    \begin{tabular}{|l|rrrrrrrrr|}
    \hline
    Inst. & \textbf{Co}  & \textbf{Cl}  & \textbf{R}  & \textbf{S}  & \textbf{W}  & \textbf{CoS}  & \textbf{ClS}  & \textbf{TCl}  &  \textbf{RCl}  \\
\hline    
    \textsf{agh-fis-spr17} & 340   & 1239  & 80    & 1641  & 16    & 8.17  & 16.2  & 117.73 & 15.92 \\
    \textsf{agh-ggis-spr17} & 272   & 1852  & 44    & 2116  & 16    & 6.98  & 29.92 & 25.2  & 7.28 \\
    \textsf{bet-fal17} & 353   & 983   & 62    & 3018  & 16    & 6.24  & 9.08  & 23.77 & 25.43 \\
    \textsf{iku-fal17} & 1206  & 2641  & 214   & 0     & 14    &   ---    &   ---    & 35.36 & 30.76 \\
    \textsf{mary-spr17} & 544   & 882   & 90    & 3666  & 16    & 2.88  & 2.9   & 13.98 & 13.57 \\
    \textsf{muni-fi-spr16} & 228   & 575   & 35    & 1543  & 15    & 6.24  & 10.06 & 16.62 & 4.82 \\
    \textsf{muni-fsps-spr17} & 226   & 561   & 44    & 865   & 19    & 7.76  & 11.6  & 20.24 & 3.15 \\
    \textsf{muni-pdf-spr16c} & 1089  & 2526  & 70    & 2938  & 13    & 8.72  & 17.35 & 59.61 & 11.82 \\
    \textsf{pu-llr-spr17} & 697   & 1001  & 75    & 27018 & 16    & 3.03  & 3.4   & 9.28  & 15.23 \\
    \textsf{tg-fal17} & 36    & 711   & 15    & 0     & 14    & ---      &  ---     & 25.86 & 4.41 \\
    \hline
    \textsf{agh-ggos-spr17} & 406   & 1144  & 84    & 2254  & 16    & 7.01  & 13.94 & 93.58 & 10.92 \\
    \textsf{agh-h-spr17} & 234   & 460   & 39    & 1988  & 16    & 2.6   & 4.18  & 236.35 & 25.47 \\
    \textsf{lums-spr18} & 313   & 487   & 73    & 0     & 20    &  ---     &  ---     & 43.86 & 27.19 \\
    \textsf{muni-fi-spr17} & 186   & 516   & 35    & 1469  & 14    & 6.22  & 10.3  & 18.92 & 5.25 \\
    \textsf{muni-fsps-spr17c} & 116   & 650   & 29    & 395   & 14    & 6.98  & 32.94 & 124.74 & 5.06 \\
    \textsf{muni-pdf-spr16} & 881   & 1515  & 83    & 3443  & 13    & 9.2   & 10.04 & 32.76 & 17.47 \\
    \textsf{nbi-spr18} & 404   & 782   & 67    & 2293  & 15    & 6.03  & 12.46 & 38.09 & 4.83 \\
    \textsf{pu-d5-spr17} & 212   & 1061  & 84    & 13497 & 15    & 1.45  & 2.46  & 11.79 & 8.77 \\
    \textsf{pu-proj-fal19} & 2839  & 8813  & 768   & 38437 & 17    & 4.71  & 6.95  & 13.43 & 9.83 \\
    \textsf{yach-fal17} & 91    & 417   & 28    & 821   & 16    & 5.07  & 13.14 & 43.98 & 4.61 \\
   \hline
    \textsf{agh-fal17} & 1363  & 5081  & 327   & 6925  & 18    & 8.7   & 20.91 & 75.55 & 10.52 \\
    \textsf{bet-spr18} & 357   & 1083  & 63    & 2921  & 16    & 6.52  & 10.46 & 23.17 & 25.15 \\
    \textsf{iku-fal18} & 1290  & 2782  & 208   & 0     & 13    &   ---    & ---   & 32.72 & 27.72 \\
    \textsf{lums-fal17} & 328   & 502   & 73    & 0     & 20    &    ---   &  ---     & 43.5  & 26.54 \\
    \textsf{mary-fal18} & 540   & 951   & 93    & 5051  & 16    & 4.16  & 4.17  & 11.37 & 15.11 \\
    \textsf{muni-fi-fal17} & 188   & 535   & 36    & 1685  & 13    & 6.59  & 10.43 & 16.3  & 4.94 \\
    \textsf{muni-fspsx-fal17} & 515   & 1623  & 33    & 1152  & 21    & 8.87  & 21.82 & 67.85 & 4.42 \\
    \textsf{muni-pdfx-fal17} & 1635  & 3717  & 86    & 5651  & 13    & 9.84  & 15.94 & 66.74 & 18.48 \\
    \textsf{pu-d9-fal19} & 1154  & 2798  & 224   & 35213 & 15    & 3.51  & 4.37  & 13.89 & 14.24 \\
    \textsf{tg-spr18} & 44    & 676   & 18    & 0     & 16    &  ---     & ---      & 23.37 & 5.67 \\
    \hline
    \end{tabular}
\caption{Features of the \itciv{} benchmark instances. Symbol definition:
\textbf{Co} (Course), \textbf{Cl} (Classes), \textbf{R} (Rooms), \textbf{S} (Students), \textbf{W} (Weeks), 
\textbf{CoS} (average Courses for Student), \textbf{ClS} (average Classes for Student), 
\textbf{TCl} (average Times of a Class), \textbf{RCl} (average Rooms of a Class).}
  \label{tab:itc2019-features}
\end{footnotesize}
\end{table}

\paragraph{File formats and repositories}

Instances are written in XML format and available from the competition
website (\url{https://www.itc2019.org}) after registering.  In
addition, the winners of ITC-2019 have implemented a preprocessing
procedure for the \itciv{} datasets \citep{HMSS19} that reduce
instances to a simplified, though still complete, form.  The reduced
\itciv{} datasets are available at \url{https://dsumsoftware.com/itc2019/}.

\paragraph{Other datasets and generators}

Up to our knowledge, there are no other instances available apart from the 
six test instances provided in the competition website.

\paragraph{Discussion}
Although the formulation mostly adhere to reality, some aspects of real-life data 
have been neglected or transformed into existing constraints
in order to make the formulation easier to model and to work on it. The most important changes 
involve the computation of the list of rooms available for a class and their individual penalties,  travel times, 
translation of distribution constraints, and student reservation.

\subsection{Other Formulations}
\label{sec:other-formulations}

We now review the additional problem formulations that provide
real-world instances that are publicly available.
Table~\ref{tab:other-formulations} shows the list of available ones,
up to our knowledge, along with the information whether the solutions
and a solution checker are available.

It is worth mentioning that there are many papers claiming that the
search method has been applied to real-world cases, but then they do
not provide the corresponding files (mainly for privacy issues). There
are also many cases in which the link for retrieving the instances is
not working anymore, typically due to authors changing affiliation.
The latter phenomenon clearly show that the strategy of posting data
in author's website does not work in the long run.  In some cases, the
link has been restored by the authors upon our specific request.

\begin{table}
\centering
  \begin{footnotesize}
    \begin{tabular}{|p{2.3cm}|r|r|c|c|l|p{2cm}|p{3.7cm}|}\hline
      Reference & Prob & \#Inst &  Sol & Check & Format & Source & link \\ \hline
      \citet{BMKL08} & \htt & 11 & $\times$ & $\times$ & text & Greece & \scriptsize \url{https://www.dropbox.com/s/rolhmd31bmrea4a/Input%20instances.zip} \\ \hline
      \citet{RuMM11} & \ctt & 50 & $\surd$ & $\surd$ & XML & Purdue (US) & \scriptsize \url{https://www.unitime.org} \\ \hline
      \citet{Mull16} & \ett & 9 & $\surd$ & $\surd$ & XML & Purdue (US) & \scriptsize \url{https://www.unitime.org} \\ \hline
      \citet{WDBC16} & \ett & 1 & $\surd$ & $\times$ & Excel & Belgium & \scriptsize \url{https://www.kuleuven.be/cv/personallinks/u0038694e.htm} \\ \hline
      \citet{SaCo17} & \htt & 34 & $\times$ & $\times$ & XML & Brazil & \scriptsize \url{https://www.gpea.uem.br/benchmark.html} \\ \hline
      \citet{LMML19} & \ctt & 8 & $\surd$ & $\surd$ & XML & Lisbon (PT) & \scriptsize \url{https://github.com/ADDALemos/MPPTimetables} \\ \hline
      \citet{BCDDST20} & \ett & 40 & $\surd$ & $\surd$ & JSON & Italy & \scriptsize \url{https://bitbucket.org/satt/examtimetablinguniuddata} \\ \hline
      \citet{GGKB21} & \ctt & 1 & $\times$ & $\times$ & Excel & Y{\i}ld{\i}z (TR) & \scriptsize \url{https://sites.google.com/view/mgguler/datasets} \\ \hline
    \end{tabular}
  \end{footnotesize}
  \caption{Other formulations and datasets. \#Inst: number of instances, Sol: solutions available ($\surd$ = Yes, $\times$ = No), Check: checker available, Source: single institution or country in case of many institutions.}
  \label{tab:other-formulations}
\end{table}

\section{State-of-the-Art Results}
\label{sec:results}

In this section, we report the results for each of the formulations
introduced in Sections~\ref{sec:uett}~---~\ref{sec:itciv}.  For each
formulation, among all results in the literature, we select
and report the ones that we consider ``state-of-the-art'', intending
with this term those that have the best scores for some instances.
However, in this selection we take into account also the running time,
thus including also results that are worse than others but obtained
with significantly shorter time.

For each contribution, we show, if available, the average and the best
scores for each instance, along with the running time (when relevant). 
Further details, such as the computing speed and the number of threads
of the machines are neglected, and can be retrieved (if reported) in
the corresponding articles.

The tables include also, when available, the best lower bound and the
best known result (upper bound), specifying also the researchers that
have found them.
In addition, the lowest best values are in italics
and the proven optimal solutions are underlined (except for perfect
solutions). Finally, top average results in the table are in
boldface. 

\subsection{Results on \uett{}}
\label{sec:res-uett} 

The state-of-the-art results on Toronto benchmarks described in
Section~\ref{sec:uett} are shown in Table~\ref{tab:uett-results}. The
last two columns report the LBs and the best UBs. The UBs are obtained
by several authors, whereas the LBs are all obtained by \cite{GDNV21}.
The letter beside each UB value indicates who are the authors: ``B"
stands for \citet{BEMP10}, ``BB" for \cite{BuBy16}, ``L"
for \cite{LFMR18} and ``BC" for \cite{BCDS21}.

We remark that there are some early results for which it is not clear
whether they were obtained on the original input data (see discussion
on Section~\ref{sec:uett}). Therefore, we decided to remove them and
to bound to fully trustworthy results only.

The proposed methods have different running times, reported in the
right-most three columns of Table~\ref{tab:uett-results}. Therefore a
completely fair comparison is not possible, given that \uett{} is
particularly sensible to the running time. In fact, longer runs
consistently produce results better than shorter ones. As a
consequence, highlighted values do not identify univocally the
``best'' contributions, as they compete with different timeouts. For
this reason, for \cite{BCDS21} we report the results of both the short
and long runs, even though the short ones are clearly inferior to the
long ones, but can be considered as competitive for the allotted time.

We also notice that all methods are metaheuristics, and there are no
approaches such as mathematical and constraint programming among the
most successful ones. As we will see in the next sections, this is not
the case for some of the other formulations (see
Sections~\ref{sec:cbctt-results},~\ref{sec:xhstt-results},
and~\ref{sec:itciv-results}).

We can see that, unfortunately, the LBs \citep{GDNV21} are not
particularly tight, leaving room for improvements.
        
\begin{sidewaystable}
  \centering
\begin{footnotesize}
  \caption{Results on Toronto benchmarks of \uett. \label{tab:uett-results}}
    \begin{tabular}{|l|rr|rr|rr|rr|rr|rr|r|r|}\hline
          & \multicolumn{2}{c|}{ \citeauthor{BCDS21}} & \multicolumn{2}{c|}{\citeauthor{BuBy08}} & \multicolumn{2}{c|}{\citeauthor{MaKK20}} & \multicolumn{2}{c|}{\citeauthor{BuBy16}}  &\multicolumn{2}{c|}{\citeauthor{LFMR18}} & \multicolumn{2}{c|}{\citeauthor{BCDS21}} & \citeauthor{GDNV21} &\\
          & \multicolumn{2}{c|}{ \citeyear{BCDS21}} & \multicolumn{2}{c|}{\citeyear{BuBy08}} & \multicolumn{2}{c|}{\citeyear{MaKK20}} & \multicolumn{2}{c|}{\citeyear{BuBy16}}  &\multicolumn{2}{c|}{\citeyear{LFMR18}} & \multicolumn{2}{c|}{\citeyear{BCDS21}} &  \citeyear{GDNV21} & \\        
    Inst. & avg & best & avg   & best   & avg   & best   & avg   & best & avg & best & avg   & best   & LB   & UB$^{\phantom{X.XX}}$  \\\hline
    \textsf{car91} &4.44&4.38&4.68&4.58&4.72&4.58&4.34&4.32&4.39&4.31& \textbf{4.27}  & \emph{4.24}  &0.0059& 4.237932$^{BC\phantom{.X}}$\\
    \textsf{car92} &3.8&3.75&3.92&3.81&3.93&3.82&3.7&3.67&3.72&3.68& \textbf{3.68}  & \emph{3.64}  &0.0079& 3.642109$^{BC\phantom{.X}}$ \\
    \textsf{ear83} &32.89&32.61&32.91&32.65&34.49&33.23&32.66&32.62&32.61&32.48& \textbf{32.60} & \emph{32.42} &18.2596& 32.420444$^{BC\phantom{.X}}$ \\
    \textsf{hec92} &10.16&10.05&10.22&10.06&11.09&10.32&10.12&10.06& \textbf{10.05} & \emph{10.03} & \textbf{10.05} & \emph{10.03} &3.8162& 10.033652$^{L.BC}$ \\
    \textsf{kfu93} &13.06&12.87&13.02&12.81&13.97&13.34&12.85& \emph{12.8}  & \textbf{12.83} & \emph{12.81} &12.88& \emph{12.81} &5.736& 12.799028$^{BC\phantom{.X}}$ \\
    \textsf{lse91} &10.09&9.92&10.14&9.86&10.62&10.24&9.84& \emph{9.78}  &9.81& \emph{9.78}  & \textbf{9.80}  & \emph{9.78}  &3.3555& 9.773661$^{BC\phantom{.X}}$ \\
    \textsf{pur93} &4.32&4.22&4.71&4.53&       &       & \textbf{3.91} & \emph{3.88}  &4.18&4.14&4.02&4&0.0014& 3.88$^{BB\phantom{.X}}$  \\
    \textsf{rye93} &8.1&7.99&8.06&7.93&10.29&9.79&7.94&7.91&7.93&7.89& \textbf{7.91}  & \emph{7.84}  &3.7868& 7.837586$^{BC\phantom{.X}}$  \\
    \textsf{sta83} &157.05& \emph{157.03} &157.05& \emph{157.03} &157.64&157.14&157.04& \emph{157.03} & \textbf{157.03} & \emph{157.03} & \textbf{157.03} & \emph{157.03} &152.0458& 156.86$^{B\phantom{.XX}}$  \\
    \textsf{tre92} &7.85&7.72&7.89&7.72&8.03&7.74&7.68&7.64&7.7&7.66& \textbf{7.66}  & \emph{7.59}  &0.8601& 7.590367$^{BC\phantom{.X}}$ \\
    \textsf{uta92} &3.13&3.05&3.26&3.16&3.22&3.13&3.01&2.98&3.04&3.01& \textbf{2.97}  & \emph{2.95}  &0.0022& 2.947193$^{BC\phantom{.X}}$\\
    \textsf{ute92} &24.82& \emph{24.76} &24.82&24.79&26.04&25.28&24.82&24.78&24.83&24.8& \textbf{24.79} & \emph{24.76} &21.5993& 24.76$^{BC\phantom{.X}}$ \\
    \textsf{yor83} &34.93&34.56&36.16&34.78&36.79&35.68&34.79&34.71&34.63&34.45& \textbf{34.57} & \emph{34.40} &19.1435& 34.404888$^{BC\phantom{.X}}$ \\\hline
    Time & \multicolumn{14}{c|}{}\\\hline
    Min   &130.8& s     &450.0& s     &       &       &4.6& h     &24& h     &26.2& h     &       & \\
    Max   &1382.0& s     &901.0& s     &       &       &5.7& h     &48& h     &52.2& h     &       &\\
    Avg   &413.1& s     &654.6& s     &1&h&5.1& h     &31.4& h     &34.7& h     &       &\\\hline
  \end{tabular}
   \end{footnotesize}
  \end{sidewaystable}

\subsection{Results on \pectt}

For the \pectt{} formulation, the results that we consider
state-of-the-art are shown in
Tables~\ref{tab:pectt-itc2002},~\ref{tab:pectt-itc2007-public}
and~\ref{tab:pectt-itc2007-hidden}, for the two datasets identified as
benchmarks in Section~\ref{sec:formulations-benchmarks}. The second
dataset is split into two tables because the first set of 16 instances and
the second one of 8 instances have been considered by different
authors, as the latter have been released at a later stage.

All results are obtained from 31 runs, using the time limit allowed by
the competition benchmark program (about 300s). For each instance, the
top average result is shown in boldface, whereas the lowest best value
is shown in italic. For the ITC-2002 benchmarks, the column UB reports
the best known value, which in this case is the lowest value in the
table, except for instance 1 for which it has been obtained by
\cite{GoKS17} with longer (five times) timeout, and instances 10 and
11 obtained by \cite{Naga18} using a method different from the most
performing one reported here.

\begin{table}[htbp]
  \centering
  \caption{Results on ITC-2002 benchmarks of \pectt.}
  \begin{footnotesize}
    \begin{tabular}{|l|r|rr|rr|rr|rr|r|}\hline
     & \citeauthor{Kost04} & \multicolumn{2}{c|}{\citeauthor{GoKS17}} & \multicolumn{2}{c|}{\citeauthor{Naga18}} & \multicolumn{2}{c|}{\citeauthor{GoKS19}} & \multicolumn{2}{c|}{\citeauthor{GKSA20}} &  \\
& \citeyear{Kost04} & \multicolumn{2}{c|}{\citeyear{GoKS17}} & \multicolumn{2}{c|}{\citeyear{Naga18}} & \multicolumn{2}{c|}{\citeyear{GoKS19}} & \multicolumn{2}{c|}{\citeyear{GKSA20}} &  \\
   Inst. & best  & avg   & best  & avg   & best  & avg   & best  & avg   & best  & UB \\\hline
   \textsf{01} & \emph{16} & 32.6  & 23    & \textbf{30.2} & \emph{16} & 37    & 26    & 36.8  & 29    & 10 \\
   \textsf{02} & \emph{2} & 13.7  & 7     & \textbf{11.4} & \emph{2} & 16.3  & 6     & 16.2  & \emph{2} & 2 \\
   \textsf{03} & \emph{17} & 36.4  & 26    & \textbf{31} & \emph{17} & 38.2  & 27    & 34.3  & 24    & 17 \\
   \textsf{04} & \emph{34} & 63.1  & 50    & \textbf{60.8} & \emph{34} & 69    & 47    & 70.7  & 46    & 34 \\
   \textsf{05} & 42    & 58.6  & 38    & 72.1  & 42    & \textbf{51.8} & \emph{36} & 55    & 43    & 36 \\
   \textsf{06} & 0 & 0.8   & 0 & 2.4   & 0 & 0.8   & 0 & \textbf{0.4} & 0 & 0 \\
   \textsf{07} & 2     & 2.6   & 0 & 8.9   & 2     & \textbf{2.4} & 0 & \textbf{2.4} & 0 & 0 \\
   \textsf{08} & 0 & \textbf{1.4} & 0 & 2     & 0 & 1.5   & 0 & 2.2   & 0 & 0 \\
   \textsf{09} & 1     & \textbf{4.6} & 0 & 5.8   & 2     & 6.4   & 0 & 6.5   & 0 & 0 \\
   \textsf{10} & \emph{21} & 40.9  & 28    & \textbf{35} & \emph{21} & 40.4  & 22    & 39.2  & 26    & 18 \\
   \textsf{11} & \emph{5} & 17.7  & 10    & \textbf{12.9} & \emph{5} & 19    & 10    & 19.7  & 9     & 4 \\
   \textsf{12} & 55    & 64.5  & 53    & 76.3  & 55    & 64.1  & 47    & \textbf{63.9} & \emph{46} & 46 \\
   \textsf{13} & \emph{31} & 53.3  & 38    & \textbf{47.1} & \emph{31} & 51    & 33    & 51.2  & 40    & 31 \\
   \textsf{14} & 11    & 12.9  & 5     & 22.3  & 11    & 13.6  & \emph{4} & \textbf{12.1} & \emph{4} & 4 \\
   \textsf{15} & 2     & \textbf{4.0} & 0 & 8.4   & 2     & 4.8   & 0 & 4.4   & 0 & 0 \\
   \textsf{16} & 0 & \textbf{0.5} & 0 & 3.4   & 0 & 2.2   & 0 & 1.6   & 0 & 0 \\
   \textsf{17} & 37    & 41.6  & 26    & 54    & 37    & \textbf{36.8} & 25    & 38.7  & \emph{24} & 24 \\
   \textsf{18} & 4     & 9.7   & \emph{2} & \textbf{9.4} & 4     & 12.5  & 3     & 11.7  & 4     & 2 \\
   \textsf{19} & \emph{7} & 24.7  & 11    & \textbf{16.4} & \emph{7} & 25.6  & 15    & 23.6  & 9     & 7 \\
   \textsf{20} & 0 & \textbf{0} & 0 & 0.5   & 0 & \textbf{0} & 0 & \textbf{0} & 0 & 0 \\\hline
     Avg     &       & 24.18 &       & 25.52 &       & 24.67 &       & 24.53 &       &  \\\hline
    \end{tabular}
    \end{footnotesize}
  \label{tab:pectt-itc2002}
\end{table}

\begin{table}[htbp]
  \centering
  \caption{Results on ITC-2007 public benchmarks of \pectt.}
  \begin{footnotesize}
    \begin{tabular}{|l|rr|rr|rr|rr|rr|}\hline
          & \multicolumn{2}{c|}{\citeauthor{MNCR08}} & \multicolumn{2}{c|}{\citeauthor{GoKS17}} & \multicolumn{2}{c|}{\citeauthor{Naga18}} & \multicolumn{2}{c|}{\citeauthor{GoKS19}} & \multicolumn{2}{c|}{\citeauthor{GKSA20}} \\
 & \multicolumn{2}{c|}{\citeyear{MNCR08}} & \multicolumn{2}{c|}{\citeyear{GoKS17}} & \multicolumn{2}{c|}{\citeyear{Naga18}} & \multicolumn{2}{c|}{\citeyear{GoKS19}} & \multicolumn{2}{c|}{\citeyear{GKSA20}} \\

   Inst. & avg   & best  & avg   & best  & avg   & best  & avg   & best  & avg   & best \\\hline
   \textsf{01} & 613   & 0     & 307.6 & 0     & \textbf{81.7} & 0     & 209.4 & 0     & 191.8 & 0 \\
   \textsf{02} & 556   & 0     & 63.4  & 0     & 48    & 0     & 10.1  & 0     & \textbf{1.7} & 0 \\
   \textsf{03} & 680   & 110   & 199.4 & 163   & \textbf{155} & \emph{55}    & 188.6 & 141   & 189.8 & 137 \\
   \textsf{04} & 580   & 53    & 328.8 & 242   & \textbf{254.1} & \emph{10}    & 320.9 & 192   & 315.5 & 24 \\
   \textsf{05} & 92    & 13    & 2.7   & 0     & \textbf{0} & 0     & 2.9   & 0     & 2.9   & 0 \\
   \textsf{06} & 212   & 0     & 33.2  & 0     & \textbf{0} & 0     & 54.7  & 0     & 37.6  & 0 \\
   \textsf{07} & 4     & 0     & 18    & 5     & \textbf{3.6} & 0     & 14.5  & 4     & 16.2  & 5 \\
   \textsf{08} & 61    & 0     & \textbf{0} & 0     & \textbf{0} & 0     & 1.6   & 0     & 5.7   & 0 \\
   \textsf{09} & 202   & 0     & 100.7 & 0     & \textbf{58.9} & 0     & 15.2  & 0     & 2.6   & 0 \\
   \textsf{10} & \textbf{4} & 0     & 65.3  & 0     & 6.4   & 0     & 30.5  & 0     & 16.3  & 0 \\
   \textsf{11} & 774   & 143   & 244.3 & 161   & \textbf{140.4} & \emph{3}     & 201.6 & 136   & 199.6 & 21 \\
   \textsf{12} & 538   & 0     & 318.2 & 0     & \textbf{33.1} & 0     & 303.5 & 0     & 258.1 & 0 \\
   \textsf{13} & 360   & 5     & 99.5  & 0     & \textbf{0} & 0     & 90.4  & 0     & 85.9  & 0 \\
   \textsf{14} & 41    & 0     & 0.2   & 0     & \textbf{0} & 0     & 25.6  & 0     & 17.8  & 0 \\
  \textsf{15} & 29    & 0     & 192   & 0     & \textbf{0} & 0     & 12.5  & 0     & 9.3   & 0 \\
   \textsf{16} & 101   & 0     & 105.8 & 10    & \textbf{1.5} & 0     & 45.8  & 0     & 40.2  & 0 \\\hline
   Avg       & 302.9 &       & 129.9 &       & 48.9  &       & 95.5  &       & 86.9  &  \\\hline
    \end{tabular}
    \end{footnotesize}
  \label{tab:pectt-itc2007-public}
\end{table}

\begin{table}[htbp]
  \centering
  \caption{Results on ITC-2007 hidden benchmarks of \pectt.}
\begin{footnotesize}
    \begin{tabular}{|l|rr|rr|rr|rr|rr|rr|rr|}\hline
         & \multicolumn{2}{c|}{Cambazard} & \multicolumn{2}{c|}{Ceschia} & \multicolumn{2}{c|}{Lewis and} & \multicolumn{2}{c|}{Goh} & \multicolumn{2}{c|}{\citeauthor{Naga18}} & \multicolumn{2}{c|}{Goh} & \multicolumn{2}{c|}{Goh} \\
&  \multicolumn{2}{c|}{et al.}  &  \multicolumn{2}{c|}{et al.} & \multicolumn{2}{c|}{Thompson} &  \multicolumn{2}{c|}{et al.} &  \multicolumn{2}{c|}{}  &  \multicolumn{2}{c|}{et al.} 
&  \multicolumn{2}{c|}{et al.} \\
                  & \multicolumn{2}{c|}{\citeyear{CHOP10}} & \multicolumn{2}{c|}{\citeyear{CeDS12}} & \multicolumn{2}{c|}{\citeyear{LeTh15}} & \multicolumn{2}{c|}{\citeyear{GoKS17}} & \multicolumn{2}{c|}{\citeyear{Naga18}} & \multicolumn{2}{c|}{\citeyear{GoKS19}} & \multicolumn{2}{c|}{\citeyear{GKSA20}} \\
    Inst. & avg & best & avg & best & avg & best & avg   & best  & avg   & best  & avg   & best  & avg   & best \\\hline
    \textsf{17}    & 4.9   & 0     & \textbf{0.0} & 0     & 0.07  & 0     & 0.8   & 0     & 0.2   & 0     & 0.5   & 0     & 0.1   & 0 \\
    \textsf{18}    & 14.1  & 0     & 41.1  & 0     & 2.16  & 0     & 12.5  & 0     & \textbf{0.5} & 0     & 7.7   & 0     & 15.5  & 0 \\
    \textsf{19}    & 2027.0 & 1824  & 951.5 & 0     & 346.08 & 0     & 516.7 & 0     & 616.8 & 0     & \textbf{11} & 0     & 79.6  & 0 \\
    \textsf{20}    & 505.0 & 445   & 700.2 & 543   & 724.54 & 557   & 650.7 & 586   & \textbf{482} & \emph{438}   & 664   & 555   & 661.5 & 579 \\
    \textsf{21}    & 27.1  & 0     & 35.9  & 5     & 32.09 & 1     & 12.5  & 0     & \textbf{0.1} & 0     & 25.7  & 0     & 14.8  & 0 \\
    \textsf{22}    & 550.8 & 29    & 19.9  & 5     & 1790.08 & 4     & 136   & 1     & 35    & 0     & \textbf{5.8} & 0     & 22.6  & 0 \\
    \textsf{23}    & \textbf{330.5} & 238   & 1707.7 & 1292  & 514.13 & 0     & 504.4 & 11    & 1083.5 & 777   & 713.6 & 56    & 531.7 & 0 \\
    \textsf{24}    & 124.2 & 21    & 105.3 & 0     & 328.18 & 18    & 192.6 & 5     & 1     & 0     & \textbf{77.5} & 0     & 102.1 & 0 \\\hline
    Avg   & 448.0 &       & 445.2 &       & 467.2 &       & 253.3 &       & 277.4 &       & 188.2 &       & 178.5 &  \\\hline
     \end{tabular}
 \end{footnotesize}
  \label{tab:pectt-itc2007-hidden}
\end{table}

We do not report the UBs in Tables~\ref{tab:pectt-itc2007-public}
and~\ref{tab:pectt-itc2007-hidden} as most of them are equal to 0 (for
instance 11 by \citealt{LeTh15}, not in the table). The only
distinctive instances are 3, 4, and 20 with UB values 55, 10 (reported
in the corresponding table), and 150 (found by \citealt{Naga18} with
another method), respectively. For ITC-2002, conversely, for many
instances, the perfect solution is still to be found.

The first comment on these tables is that all best results have been
found by local search methods, namely Tabu Search \citep{Naga18} and
Simulated Annealing \citep{GoKS17,GoKS19,GKSA20}. In general, best
results are obtained by \citet{Naga18}, that uses a composite
neighborhood and \emph{elite candidate} rules to reduce the
computational cost of the full neighborhood exploration prescribed by
Tabu Search. Good results are obtained also by \citeauthor{GKSA20},
mainly using random move selection.

\citet{GoKS19} report also the results for longer running times (i.e.,
five times longer), which are not shown here. Unsurprisingly, both the
best and average cost are remarkably improved when the execution time
is extended.

It is worth noticing that the fact that all instances have a perfect
(zero cost) solution might bias the search methods toward certain
specific strategies. For example, the objective that penalizes all
lectures in the last period of the day might be exploited, by removing
such periods completely from the search space.

\subsection{Results on \cbctt}
\label{sec:cbctt-results}

\begin{table}[htbp]
  \centering
  \caption{Results on ITC-2007 benchmarks of \cbctt.}
 \begin{footnotesize}
    \begin{tabular}{|l|rr|rr|r|rr|}\hline
          & \multicolumn{2}{c|}{\citeauthor{AbTu12}} & \multicolumn{2}{c|}{\citeauthor{KiHS17}} & \multicolumn{1}{c|}{\citeauthor{LiSS18}} &       &  \\
          & \multicolumn{2}{c|}{\citeyear{AbTu12}} & \multicolumn{2}{c|}{\citeyear{KiHS17}} & \multicolumn{1}{c|}{\citeyear{LiSS18}} &       &  \\
          & avg   & best  & avg   & best  & avg   & \multicolumn{1}{c}{LB}    & \multicolumn{1}{c|}{UB} \\\hline
   \textsf{comp01} & \textbf{5.00} & \emph{\underline{5}} & \textbf{5.0} & \emph{\underline{5}} & 12.0  & $5^{C,B1\phantom{,}}$ & $5^{\ast\phantom{,,L1}}$ \\
   \textsf{comp02} & \textbf{36.36} & \emph{26}    & 41.5  & 34    & 49.5  & $24^{B2\phantom{,,X}}$ & $24^{A\phantom{,L1}}$ \\
   \textsf{comp03} & 74.36 & 70    & \textbf{71.7} & \emph{68}    & 74.5  & $58^{B3\phantom{,,X}}$ & $64^{K\phantom{,L1}}$ \\
   \textsf{comp04} & 38.45 & \emph{\underline{35}} & \textbf{35.1} & \emph{\underline{35}} & 38.5  & $35^{\ast\phantom{,,XX}}$ & $35^{\ast\phantom{,,L1}}$ \\
   \textsf{comp05} & 314.45 & 295   & \textbf{305.2} & \emph{294}   & 373.5 & $247^{B3\phantom{,,X}}$ & $284^{K\phantom{{,L1}}}$ \\
   \textsf{comp06} & \textbf{45.27} & \emph{30}    & 47.8  & 41    & 58.3  & $27^{A\phantom{,XX}}$ & $27^{A\phantom{,,L1}}$ \\
   \textsf{comp07} & \textbf{12.00} & \emph{7}     & 14.5  & 10    & 35.0  & $6^{\ast\phantom{,,XX}}$ & $6^{A\phantom{,,L1}}$ \\
   \textsf{comp08} & \textbf{40.82} & \emph{\underline{37}} & 41.0  & 39    & 49.7  & $37^{\ast\phantom{,,XX}}$ & $37^{A\phantom{,,L1}}$ \\
   \textsf{comp09} & 108.36 & 102   & 102.8 & \emph{100}   & \textbf{100.5} & $96^{B2\phantom{,,X}}$ & $96^{L1\phantom{,L1}}$ \\
   \textsf{comp10} & \textbf{8.36} & \emph{5}     & 14.3  & 7     & 25.7  & $4^{\ast\phantom{,,XX}}$ & $4^{A\phantom{,,L1}}$ \\
   \textsf{comp11} & 0.00  & \emph{\underline{0}} & 0.0   & \emph{\underline{0}}     & 6.5   & $0^{\ast\phantom{,,XX}}$ & $0^{\ast\phantom{,,L1}}$ \\
   \textsf{comp12} & 320.27 & 315   & \textbf{319.4} & \emph{306}   & 360.7 & $248^{B3\phantom{,,X}}$ & $294^{K\phantom{,L1}}$ \\
   \textsf{comp13} & 64.27 & \emph{\underline{59}}    & \textbf{60.7} & \emph{\underline{59}} & 69.0  & $59^{\ast\phantom{,,XX}}$ & $59^{A\phantom{,L1}}$ \\
   \textsf{comp14} & 64.36 & 61    & \textbf{54.1} & \emph{\underline{51}} & 56.9  & $51^{\ast\phantom{,,XX}}$ & $51^{A,L1}$ \\
   \textsf{comp15} & 72.73 & 69    & \textbf{72.1} & \emph{66}    & 74.5  &$58^{B3\phantom{,,X}}$ & $62^{K\phantom{,L1}}$ \\
   \textsf{comp16} & \textbf{23.73} & \emph{\underline{18}} & 33.8  & 26    & 37.1  & $18^{A,B2\phantom{,}}$ & $18^{A\phantom{,L1}}$ \\
   \textsf{comp17} & 76.36 & \emph{60}    & \textbf{75.7} & 67    & 86.1  & $56^{A,B3\phantom{,}}$ & $56^{A\phantom{,L1}}$ \\
   \textsf{comp18} & 75.64 & 69    & \textbf{66.9} & \emph{64}    & 72.9  & $61^{L2\phantom{,,X}}$ & $61^{K\phantom{,L1}}$ \\
   \textsf{comp19} & 66.82 & \emph{\underline{57}} & \textbf{62.6} & 59    & 64.8  & $57^{B2\phantom{,,X}}$ & $57^{L1\phantom{,L1}}$ \\
   \textsf{comp20} & \textbf{13.45} & \emph{7}     & 27.2  & 19    & 34.3  & $4^{\ast\phantom{;,XX}}$ & $4^{A\phantom{,,L1}}$ \\
   \textsf{comp21} & 100.73 & \emph{86} & \textbf{97.0} & 93    & 103.8 & $74^{B2,L2}$ & $74^{P\phantom{,L1}}$ \\\hline
    Avg      & 74.37 & 67.29 & \textbf{73.73} & 68.71 & 84.94 &       &  \\\hline
    \end{tabular}
    \end{footnotesize}
  \label{tab:results-cbctt}
\end{table}

Table~\ref{tab:results-cbctt} shows the best results for \cbctt
benchmarks obtained using the timeout fixed for the ITC-2007 dataset
(300-500 seconds depending on the CPU). Longer runs, which
unsurprisingly obtain better results, are not considered here
\cite[see][]{LuHa09,AsNi14,SCXW21}. We take them into account only for
establishing the LBs and UBs, which are shown in the last two columns
of the table. In particular, the LBs are obtained with a running time
up to 40 times the ITC-2007 timeout.

Besides each best-known lower and upper bound values, we report a
letter that indicates who are the authors\footnote{We note that the
  UBs and LBs in Table 4 of \cite{LiSS18} (column Best, including the
  numbers in parentheses) are actually wrong, as they refer to the
  formulation UD1 instead of UD2 considered in that paper (and here)}:
``A" stands for \cite{AsNi14}, ``B1" for \cite{BMPR10b}, ``B2" for
\cite{BaDD19}, ``B3" for \cite{BaSS19}, ``C" for \cite{CCRT13}, ``K"
for \cite{KiHS17}, ``L1" for \cite{LuHa09}, ``L2" for Gerard Lach,
``P" for \cite{Phil15}.  If the same value was found by many different
authors, we marked it with the symbol $\ast$.

We see from Table~\ref{tab:results-cbctt} that almost all current best
results are obtained by two contributions, namely \cite{AbTu12} and
\cite{KiHS17}, who both proposed metaheuristic methods using Adaptive
Large Neighborhood operators.  \citeauthor{AbTu12} implemented a
Genetic Algorithm hybridized with Tabu Search employing large
neighborhood operators, whose sequence of employment follows a ``best"
selection strategy, based on previous knowledge about the successful
percentage of each neighborhood structure on each instance.
\cite{KiHS17} presented an Adaptive Large Neighborhood Search
algorithm embedded in a Simulated Annealing framework, incorporating
several destroy and repair operators, whose selection probability is
dynamically biased towards the best-performing ones.  Quite a few
other papers have produced results that were state-of-the-art at the
time of their publication, including
\citep{Mull08,LuHa09,ATMM10,BCDSU16}.

As remarked by \citet[Table 6]{BaSS19}, all benchmark instances but 3
are currently solved to optimality\footnote{In the paper they are
  actually 4, but as mentioned above, \texttt{comp03} and
  \texttt{comp15} are identical in this formulation.}.
In our opinion, the fact that the optimal value has been found does
not undermine the benchmarking role of these instances, which are
still challenging for medium-short timeouts. Nonetheless, there are
other public instances that are already available (on \opthub) that
could come up beside the current ones, in order to create a larger,
more comprehensive benchmark set (see Section~\ref{sec:cb-ctt}).

It is worth noticing that this is the only one among our six standard
formulations for which there has been a lot of research for finding
the best lower bounds.

\subsection{Results on \itcii}

The state-of-the-art results for \itcii{} using the ITC-2007 timeout
are shown in Table~\ref{tab:results-itcii}. First of all, we notice
that the best results are obtained mainly by Bikov and coworkers. They
use innovative local search algorithms, such as Late Acceptance and
Step Counting Hill Climbing, applied to complex neighborhood
structures (such as Kempe chains).

Research on this problem is still active and more recent results are
available \cite[see, e.g.,][]{BaSU17,LeMR19,LeMR21}; however, they do
not outperform the previous ones shown in
Table~\ref{tab:results-itcii}.

\begin{table}[htbp]
  \centering
  \caption{Results on ITC-2007 benchmarks of \itcii.}
  \begin{footnotesize}
    \begin{tabular}{|l|rr|r|r|r|r|}\hline
      & \multicolumn{2}{c|}{\citeauthor{BuBy16}} & \citeauthor{ByPe16} & \citeauthor{BuBy17} & \citeauthor{GGAK10} & \citeauthor{ArBM19} \\
      & \multicolumn{2}{c|}{\citeyear{BuBy16}} & \citeyear{ByPe16} & \citeyear{BuBy17} & \citeyear{GGAK10} & \citeyear{ArBM19} \\
      Inst.      & avg   & best     & best  & avg   & best  & LB \\\hline
      \textsf{1}     & 3792.5& 3691   & \emph{3647} & \textbf{3787} & 4128  & ---\phantom{*} \\
      \textsf{2}     & \textbf{393.1}& 385    & 385 & 402   & \emph{380} & 10\phantom{*} \\
      \textsf{3}     & 7611.8& \emph{7359}  & 7487  & \textbf{7378} & 7769  & 670\phantom{*} \\
      \textsf{4}     & \textbf{12100.4}& \emph{11329}  & 11779 & 13278 & 13103 & 1620\phantom{*} \\
      \textsf{5}     & 2512.9  & 2482  & \emph{2447}   & \textbf{2491} & 2513  &  ---\phantom{*} \\
      \textsf{6}     & 25491.5 & 25265  & \emph{25210} & \textbf{25461} & 25330 & 22875* \\ 
      \textsf{7}     & 3755.1 & 3608   & \emph{3563} & \textbf{3589} & \emph{3537} & ---\phantom{*} \\
      \textsf{8}     & 6949.9 & 6818   & \emph{6614} & \textbf{6701} & 7087  &  1250*\\ 
      \textsf{9}     & \textbf{930} & \emph{902}  & 924   & 997   & 913   &  ---\phantom{*} \\
      \textsf{10}    & \textbf{12975.7} & \emph{12900}  & 12931 & 13013 & 13053 & 0\phantom{*} \\
      \textsf{11}    & 23931.7  & \emph{22875} & 23784 & \textbf{22959} & 24369 & 3970\phantom{*} \\
      \textsf{12}    & \textbf{5176.3} & 5107   & 5097 & 5234  & \emph{5095} & 2030\phantom{*} \\\hline
    \end{tabular}
  \end{footnotesize}
  \label{tab:results-itcii}
\end{table}

The lower bounds are obtained by \cite{ArBM19} by considering only a
subset of the soft constraints. In detail, they consider the spacing
soft constraints, namely $(i)$ and $(ii)$ mentioned in
Section~\ref{sec:itcii}, and compute the number of violations induced
by the largest clique in the corresponding graph. As shown in
Table~\ref{tab:results-itcii}, for some instances the method does not
produce any result as the largest clique is not big enough to
contribute any violation. For instances \textsf{6} and \textsf{8},
marked with an *, we add to the LB computed by \citeauthor{ArBM19}
(whose original values were 2600 and 0, respectively) the fixed cost
of constraints $(iii)$ and $(v)$ due to the fact that there are not
enough periods to satisfy them (see Section~\ref{sec:itcii} for the
detailed explanation).

\subsection{Results on \xhstt}
\label{sec:xhstt-results}

\begin{table}[htbp]
  \centering
  \caption{Results on the XHSTT-2014 benchmarks of \xhstt.}
 \begin{footnotesize}
    \begin{tabular}{|l|r|r|r||r|r|r|r|}\hline      
     & Demirovi\'c & Demirovi\'c  &Teixeira& Fonseca & Kheiri and &    &  \\
     & and Musliu & and Stuckey & et al. & et al. & Keedwell &    &  \\   
    
    & \citeyear{DeMu17} & \citeyear{DeSt18} & \citeyear{TSSC19} & \citeyear{FSCS17} & \citeyear{KhKe17} &    &  \\  
    
     Inst. & avg     & avg    & avg   & z  &    best   & LB\phantom{X,X}    & UB\phantom{X} \\\hline
    \textsf{AU-BG-98} &       &       & (3, 514) &       & 493   & 0$^{\phantom{X,X}}$     & 128$^{G}$\\
    \textsf{AU-SA-96} &       &       & (16, 91) & \emph{\underline{0}} & 2     & 0$^{\phantom{X,X}}$     & 0$^{G}$ \\
    \textsf{AU-TE-99} &       &       & (7, 13) & \emph{\underline{20}} & 61    & 20$^{G\phantom{,X}}$    & 20$^{G}$ \\
    \textsf{BR-SA-00} & \emph{\underline{5}} & \emph{\underline{5}} &       &       & 10    & 5$^{L,D}$     & 5$^{L}$ \\
    \textsf{BR-SM-00} & 61.4  & 88    & 100   &       & (2, 117) & 51$^{L,D}$  & 51$^{L}$ \\
    \textsf{BR-SN-00} & 50.6  & 66    & 170   &       & 101   & 35$^{D\phantom{,X}}$    & 35$^{D}$ \\
    \textsf{DK-FG-12} &       &       &       & 1300  & 1522  & 412$^{G\phantom{,X}}$   & 1263$^{G}$ \\
    \textsf{DK-HG-12} &       &       &       & (12, 2356) & (12, 2628) & 7$^{G\phantom{,X}}$     & (12, 2330)$^{G}$ \\
    \textsf{DK-VG-09} &       &       &       & (2, 2329) & (2, 2720) & (2, 0)$^{G\phantom{,X}}$ & (2, 2323)$^{G}$ \\
    \textsf{ES-SS-08} &       &       &       & \emph{335} & 517   & 334$^{L,G}$   & 335$^{L}$ \\
    \textsf{FI-PB-98} & 54.6  & 9     &       &       & 8     & 0\phantom{X,X}     & 0$^{\phantom{X}}$ \\
    \textsf{FI-WP-06} & 9.8   & 4     & 1     &       & 7     & 0\phantom{X,X}     & 0$^{G}$ \\
    \textsf{FI-MP-06} & 95.2  & 90    & 93    &       & 89    & 77$^{L,V2}$    & 77$^{G}$ \\
    \textsf{GR-PA-08} & 5     & 7     &       &       & 4     & 3$^{L\phantom{,X}}$    & 3$^{G}$ \\
    \textsf{IT-I4-96} & 35    &       &       &       & 34    & 27$^{L\phantom{,X}}$    & 27$^{G}$ \\
    \textsf{KS-PR-11} &       &       &       &       & 3     & 0\phantom{X,X}     & 0$^{D2}$ \\
    \textsf{NL-KP-03} &       &       & 1383  & \emph{\underline{199}} & 466   & 0\phantom{X,X}     & 199$^{G}$ \\
    \textsf{NL-KP-05} &       &       & 1056  & 433   & 811   & 89$^{V2,G}$    & 425$^{G}$ \\
    \textsf{NL-KP-09} &       &       &       & \emph{1620} & (2, 7495) & 180$^{G\phantom{,X}}$   & 1620$^{G}$ \\
    \textsf{UK-SP-06} &       &       &       & (5, 4014) & (19, 1294) & 0\phantom{X,X}     & (4, 1708)$^{S}$ \\
    \textsf{US-WS-09} &       &       &       & 103   & 512   & 101$^{G\phantom{,X}}$   & 101$^{G}$ \\
    \textsf{ZA-LW-09} &       & \emph{\underline{0}} &       &       & 52    & 0\phantom{X,X}     & 0$^{V}$ \\
    \textsf{ZA-WD-09} &       &       &       &       & (9, 0) & 0\phantom{X,X}     & 0$^{L}$ \\
    \hline
    \end{tabular}
  \end{footnotesize}
  \label{tab:results-xhstt}
\end{table}

Table~\ref{tab:results-xhstt} reports the state-of-the-art results for
the benchmark instances of \xhstt. As mentioned in Section~\ref{sec:xhstt}, two
instances have been eliminated due to the fact that they are too easy.

The first three columns report average results obtained within the
competition timeout (1000 secs), whereas \citeauthor{FSCS17} and
\citeauthor{KhKe17} did not impose a time limit.  Besides each
best-known lower and upper bound values, we report a letter that
indicates who are the authors: ``G" stands for the UFOP-GOAL team
(\citeauthor*{FoSC16}), ``L" for the Lectio team
(\citeauthor*{KrSS15}), ``V" for the VAGO team (Valouxis,
Gogos, Daskalaki, Alefragis, Goulas, and Housos), ``D" for
{\'A}.~P.~Dorneles, ``V2" for M.~de Vos, ``D2" for
\citeauthor{DeMu17}, and ``S" for Skolaris (M.~Klemsa).

We first notice that most authors have not considered all instances.
For example, \cite{FSCS17} omit instances whose optimal solution is
already known and proven.

Although the ITC-2011 competition was dominated by metaheuristic
methods, recently exact methods based on integer programming
\citep{KrSS15,FSCS17,DoAB17}, maxSAT \citep{DeMu17} and constraint
programming \citep{DeSt18} have proven to be very effective for
\xhstt{}. Indeed, differently from the formulations of
Sections~\ref{sec:uett}--\ref{sec:itcii}, for \xhstt{} it ended up
being customary to use IP techniques and to evaluate the performance
of a solution methods without time limit.  Indeed, its best known
solutions and lower bounds are updated/improved by the community on
the \xhstt{} website. An up-to-date categorization of the different
solution methods applied to \htt{} (including \xhstt) is presented by
\citet{TGKS21}.

As mentioned in Section~\ref{sec:xhstt}, the previous versions of the
\xhstt{} archive are deprecated, and thus we do not include them in
the benchmarks. However, one of them, namely the hidden dataset of
ITC-2011, due to its popularity given by the competition, has been
used as testbed by many authors. In particular, there are interesting
results by \citet{KrSS15}, \citet{FoSC16}, \citet{DeMu17}, and
\citet{TSSC19}.  In addition, LBs have been found by \cite{DoAB17}.

\subsection{Results on \itciv}
\label{sec:itciv-results}

\begin{table}[htbp]
  \centering
  \caption{Results on ITC-2019 benchmarks of \itciv.}
  \begin{footnotesize}
    \begin{tabular}{|l|r|r|}\hline
          & DSU Team    &        \\
    Inst. & LB    & UB$^{\phantom{X}}$      \\\hline
    \textsf{agh-fis-spr17} & 1336  & 3039$^D$   \\
    \textsf{agh-ggis-spr17} & 23164 & 34285$^D$  \\
    \textsf{bet-fal17} & 89278 & 289965$^D$  \\
    \textsf{iku-fal17} & 18001 & 18968$^D$  \\
    \textsf{mary-spr17} & 14359 & 14910$^D$  \\
    \textsf{muni-fi-spr16} & 3556  & 3756$^D$   \\
    \textsf{muni-fsps-spr17} & 868   & \underline{868}$^D$    \\
    \textsf{muni-pdf-spr16c} & 14279 & 33724$^D$  \\
    \textsf{pu-llr-spr17} & 10038 & \underline{10038}$^D$  \\
    \textsf{tg-fal17} & 4215  & \underline{4215}$^U$   \\ \hline
    \textsf{agh-ggos-spr17} & 1844  & 2864$^D$   \\
    \textsf{agh-h-spr17} & 8945  & 21559$^D$  \\
    \textsf{lums-spr18} & 24    & 95$^D$     \\
    \textsf{muni-fi-spr17} & 2500  & 3825$^D$   \\
    \textsf{muni-fsps-spr17c} & 1361  & 2596$^D$   \\
    \textsf{muni-pdf-spr16} & 13008 & 17208$^D$  \\
    \textsf{nbi-spr18} & 18014 & \underline{18014}$^D$  \\
    \textsf{pu-d5-spr17} & 6981  & 15204$^M$  \\
    \textsf{pu-proj-fal19} & 54972 & 117425$^M$  \\
    \textsf{yach-fal17} & 516   & 1074$^M$   \\ \hline
    \textsf{agh-fal17} & 5728  & 118038$^M$  \\
    \textsf{bet-spr18} & 63444 & 348524$^D$  \\
    \textsf{iku-spr18} & 25781 & 25863$^D$  \\
    \textsf{lums-fal17} & 254   & 349$^D$    \\
    \textsf{mary-fal18} & 3496  & 4331$^D$   \\
    \textsf{muni-fi-fal17} & 1890  & 2999$^D$   \\
    \textsf{muni-fspsx-fal17} & 7747  & 10123$^M$  \\
    \textsf{muni-pdfx-fal17} & 26711 & 98373$^M$  \\
    \textsf{pu-d9-fal19} & 28000 & 39942$^D$  \\
    \textsf{tg-spr18} & 12704 & \underline{12704}$^D$  \\\hline
    \end{tabular}
 \end{footnotesize}
  \label{tab:itciv-results}
\end{table}

The competition finished in 2020, so the problem is rather new, and the only
published results are those of the competition. Differently from
previous competitions, the goal of ITC-2019 was to find all-time-best
solutions to all competition instances, without time limits or
technology restrictions.  As a consequence, in
Table~\ref{tab:itciv-results} we reported only UBs, whose solutions
are collected (and continuously updated) on the competition website
(\url{https://www.itc2019.org/}).

The competition was won by the DSU team \citep{HMSS19} who devised a
Fix-and-Optimize matheuristic, which was able to find all best
solutions except for one instance (\textsf{agh-fal17}). In addition,
the DSU teams maintains a website
(\url{https://dsumsoftware.com/itc2019/}) reporting their current best
results and the lower bounds (showed on
Table~\ref{tab:itciv-results}). The second place was obtained by
\cite{RTRH19} who modeled the problem as MIP enhanced with some
preprocessing techniques that improve its efficiency. The third place
was occupied by \cite{GaSy19} who presented a Simulated Annealing
algorithm\footnote{Their source code is available at
  \url{https://github.com/edongashi/itc-2019}}.

The letter beside each UB value in Table~\ref{tab:itciv-results}
indicate the authors: ``D" stands for the DSUM team
(\citeauthor*{HMSS19}), ``U" for the UFOP team (M, A. Pires, H.
Gambini Santos, T. A.M. Toffolo), and ``M"  for \citet{Mull20}.

\section{Conclusions and Future Directions}
\label{sec:conclusions}
\newcounter{Conclusions}

The quest for formulations and benchmarks carried out for this survey
has brought out various aspects of the current practice in timetabling
research. We summarize here our observations, and we split them in
three groups regarding the standard formulations, the specific
formulations, and the solution techniques, respectively. In our
opinion, these observation can serve as starting points for future
research directions.

Key observations about standard formulations:

\begin{enumerate}[A.]
\item Most of the standard formulations arose from timetabling
  competitions, which have given the necessary initial boost in terms
  of infrastructure and promotion. 
\item For some of the standard formulations, the benchmark instances
  are not challenging anymore, as they are too easily solved to
  optimality. Others, on the contrary, are still very challenging
  after more than 20 years from their publication.
\item There is a clear trend in the timetabling community to move
  toward rich formulations, getting rid of strong simplifications. In
  our opinion, this is a positive trend, but should be paired with the
  maintenance and renewal of the simple formulations, that could still
  serve as better testbeds for comparisons.
\setcounter{Conclusions}{\value{enumi}}
\end{enumerate}

Moving to the contributions introducing specific formulations, we have
the following observations:

\begin{enumerate}[A.] 
\setcounter{enumi}{\value{Conclusions}}
\item Many of the papers discussing original formulations do not
  provide publicly available data. For others, the original repository
  has become inaccessible after some time from the publication of the
  paper. Finally, in other cases, the file formats are too cumbersome
  and not sufficiently documented, to be easily usable for other
  researchers.
\item Most formulations are too specific for the particular case at
  hand without consideration of wider application, so that it is
  difficult to gain general insights from the papers. In addition, in
  some cases the precise formulation is not completely explained, so
  that it is not possible for other researchers to replicate the same
  model and to obtain comparable results.
\item For most formulations, the solutions are not made available, and
  thus the results in the papers could not be validated. In addition,
  the source code of the search method is very rarely available, so
  that the experiments cannot be replicated.
  \setcounter{Conclusions}{\value{enumi}}
\end{enumerate}

Regarding the comparison of solution techniques, we make the following
observations:

\begin{enumerate}[A.]
\setcounter{enumi}{\value{Conclusions}}
\item There is a need for the clear definition of the competition
  grounds, in terms of running time, statistical significance, computing
  architecture, usable technology, commercial licenses, and other
  issues. In the formulations coming form the competitions, the ground
  has been set by the official competition rules, which however might
  need to be refined and extended in order to do not harness future
  research.
\item The results of Section~\ref{sec:results} clearly show that both
  exact and (meta)heuristic techniques have their role and their
  chance to emerge, depending on the specific formulation and the
  competition ground.
\item There is a need for new benchmarks that could take over for the
  ones that turned out to be too easy for state-of-the-art techniques.
\item There is also need for more instances that could be used for the
  statistically-principled tuning of the solution methods, letting the
  benchmarks to be used only for the validation phase (avoiding
  overtuning). To this aim, the use of high quality generators could also help, 
  as them could provide an unlimited number of instances.
\end{enumerate}

All above points together highlight the need for the development of
research infrastructures in terms of common formulations, robust file
formats, long-term web repositories with instances and solutions, generators, and
solution checkers. The implementation of a wholesome and robust
infrastructure of this type is clearly too expensive in terms of human
effort to be left to the initiative of single research groups.
Therefore, there is the need for coordinated community-level actions,
in order to develop an infrastructure and, at the same time, create
the necessary consensus upon its adoption. In our opinion, to this
aim, the organization of future timetabling competitions could still
be the right key to pursuit this task.

Another point that emerged from our analysis is the issue of the
reproducibility and trustworthiness of results. In fact, the risk of
reporting false results has emerged significantly, though mainly in
the early times of the timetabling research. In any case, it is still
important that data is available for both inspection and future
comparisons. This is indeed a general issue that is ubiquitous in many
research areas, as journals currently push for publication of data
along with the papers.

We are trying to give our contribution for solving these issues by the
development of the web application \opthub, which provides a common
platform able to host new problems with their instances and solutions.
Solutions in \opthub{} are immediately validated and made available to
the community.

\opthub{} is an ongoing project, and hopefully will be extended
significantly in future releases. The main future feature will be
include in a new version is the possibility to upload the software and
to run it (also on behalf of other researchers). Hopefully, this
option will allow the community to make fairer comparisons and
statistical analyses on the behavior of the solution code.

\subsection*{Acknowledgments}
We wish to thank all the people from the timetabling community that have helped us by 
pointing out to us articles, datasets, and results. We will name 
all of them in the acknowledgments of the final version of this article.

\bibliographystyle{plainnat}
\bibliography{edutt}

\end{document}